\documentclass{article}

\PassOptionsToPackage{numbers, compress}{natbib}

\usepackage[preprint]{neurips_2023}

\usepackage[utf8]{inputenc} 
\usepackage[T1]{fontenc}    
\usepackage{hyperref}       
\usepackage{url}            
\usepackage{booktabs}       
\usepackage{amsfonts}       
\usepackage{nicefrac}       
\usepackage{microtype}      
\usepackage{xcolor}         

\usepackage[acronym]{glossaries} 
\usepackage{pifont}              
\usepackage{paralist}            
\usepackage{comment}             
\usepackage{mathtools}           
\usepackage{multirow}            
\usepackage{wrapfig}             
\usepackage{caption}             

\title{SOBER: Highly Parallel Bayesian Optimization and Bayesian Quadrature over Discrete and Mixed Spaces}

\author{
 Masaki Adachi, Saad Hamid, \textbf{Martin Jørgensen},  \textbf{Michael A. Osborne}\\
  Machine Learning Reserach Group, University of Oxford,\\
  \texttt{\texttt{\{masaki, saad, martinj, mosb\}@robots.ox.ac.uk}} \\
   \And
 Satoshi Hayakawa, Harald Oberhauser\\
  Mathematical Institute, University of Oxford,\\
  \texttt{\{hayakawa,oberhauser\}@maths.ox.ac.uk}\\
}

\begin{document}
\maketitle
\vspace{-4mm}

\begin{abstract}
Batch Bayesian optimisation and Bayesian quadrature have been shown to be sample-efficient methods of performing optimisation and quadrature where expensive-to-evaluate objective functions can be queried in parallel.
However, current methods do not scale to large batch sizes — a frequent desideratum in practice (e.g. drug discovery or simulation-based inference).
We present a novel algorithm, SOBER, which permits scalable and diversified batch global optimisation and quadrature with arbitrary acquisition functions and kernels over discrete and mixed spaces.
The key to our approach is to reformulate batch selection for global optimisation as a quadrature problem, which relaxes acquisition function maximisation (non-convex) to kernel recombination (convex). Bridging global optimisation and quadrature can efficiently solve both tasks by balancing the merits of exploitative Bayesian optimisation and explorative Bayesian quadrature.
We show that SOBER outperforms 11 competitive baselines on 12 synthetic and diverse real-world tasks.\footnote{Code: \url{https://github.com/ma921/SOBER}}\looseness=-1
\end{abstract}

\newacronym{bo}{BO}{Bayesian optimisation}
\newacronym{gp}{GP}{Gaussian process}
\newacronym{af}{AF}{acquisition function}
\newacronym{dpp}{DPP}{determinantal point process}
\newacronym{ts}{TS}{Thompson sampling}
\newacronym{sober}{SOBER}{solving optimisation as Bayesian estimation via recombination}
\newacronym{bq}{BQ}{Bayesian quadrature}
\newacronym{lp}{LP}{local penalisation}
\newacronym{ucb}{UCB}{upper confidence bound}
\newacronym{fbgp}{FBGP}{fully Bayesian GP}
\newacronym{basq}{BASQ}{Bayesian alternately subsampled quadrature}
\newacronym{rchq}{RCHQ}{random convex hull quadrature}
\newacronym{kq}{KQ}{kernel quadrature}
\newacronym{rkhs}{RKHS}{reproducing kernel Hilbert space}
\newacronym{mc}{MC}{Monte Carlo}
\newacronym{qmc}{qMC}{quasi-Monte Carlo}
\newacronym{mcmc}{MCMC}{Markov chain Monte Carlo}
\newacronym{kde}{KDE}{kernel density estimation}
\newacronym{lfi}{LFI}{likelihood-free inference}
\newacronym{abc}{ABC}{approximate Bayesian computation}
\newacronym{mes}{MES}{max-value entropy search}
\newacronym{bqbc}{B-QBC}{Bayesian query-by-committee}
\newacronym{qbmgp}{QB-MGP}{query by mixture of Gaussian process}
\newacronym{cdf}{CDF}{cumulative density function}
\newacronym{pdf}{PDF}{probability density function}
\newacronym{map}{MAP}{maximum a posteriori}
\newacronym{mle}{MLE}{maximum likelihood estimation}
\newacronym{ns}{NS}{nested sampling}
\newacronym{mvn}{MVN}{multivariate normal}
\newacronym{rmse}{RMSE}{root-mean-squared error}
\newacronym{vae}{VAE}{variational autoencoder}
\newacronym{wkde}{WKDE}{weighted kernel density estimation}
\newacronym{mmd}{MMD}{maximum mean discrepancy}
\newacronym{svd}{SVD}{singular value decomposition}
\newacronym{skq}{SKQ}{subsample-based kernel quadrature}
\newacronym{sr}{SR}{simple regret}
\newacronym{br}{BR}{Bayesian regret}
\newacronym{mv}{MV}{mean variance}
\newacronym{md}{MD}{mean distance}
\newacronym{pi}{PI}{probability of improvement}
\newacronym{ei}{EI}{expected improvement}
\newacronym{qd}{QD}{quadrature distillation}

\section{Introduction}
\begin{wrapfigure}[20]{r}{0.5\textwidth}
\vspace{-12mm}
\centering
\includegraphics[width=0.40\textwidth]{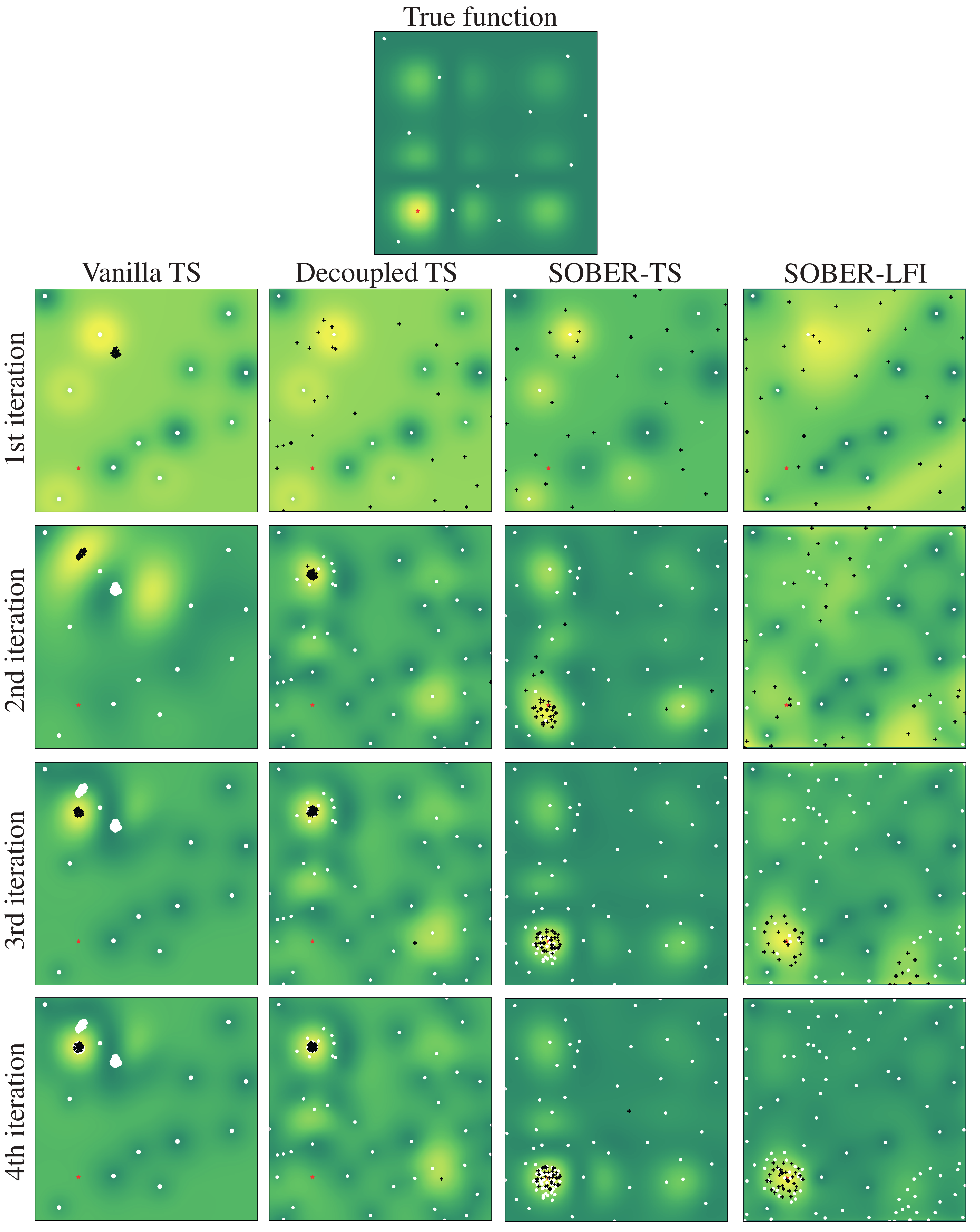}
\caption{
Thompson sampling is under-explorative. The white, black, and red points are the observed, querying, and true maximum locations.}
\label{fig:demo}
\end{wrapfigure}
\gls{bo} is a sample-efficient model-based global optimiser.
\gls{bo} typically use a \gls{gp}, whose predictive mean and variance guide the optimiser where to evaluate next by maximising the \gls{af}.
Flexibility and superb sample-efficiency enable a range of expensive-to-evaluate applications, e.g. drug discovery \cite{gomez2018automatic, griffiths2022gauche}, materials \cite{adachi2021high}, and hyperparameter optimisation \cite{feurer2015efficient, wu2020practical}.
Batch \gls{bo} offers faster convergence by querying multiple locations at once (batch acquisition). 
\gls{bq}, a model-based blackbox integration method akin to \gls{bo}, widely applied in simulation-based inference. The task is to estimate both the posterior \textit{and} marginal likelihood (also called evidence) of blackbox functions (simulators), applied to science (e.g. astrophysics \cite{osborne2012bayesian, osborne2012active}, batteries \cite{adachi2022bayesian}).

\begin{table}
  \captionsetup{skip=0pt}
  \caption{Comparing batch BO algorithms.
  The scalability to large batch sizes is defined by the computational complexity smaller or equivalent complexity to Thompson sampling (TS) (See Supp. B). Comb. explos. refers to combinatorial explosion when applied to discrete inputs with large batch size.
  The sparsity means more diversified samples than random Monte Carlo. $x$, $f$, and $h$ spaces refer to the input, function, and hyperparameter spaces. The black-box evidence represents the ability to estimate the integral of the black-box function over input space $x$, equivalent to the task of \gls{bq}.}
  \label{tab:comparison}
  \centering
  \small
  \begin{center}
  \begin{tabular}{lllllllll}
    \toprule
    Batch BOs &
    \begin{tabular}{@{}l@{}}large \\ batch\end{tabular} &
    \begin{tabular}{@{}l@{}}mixed \\ space\end{tabular} &
    \begin{tabular}{@{}l@{}}No comb. \\ explos. \end{tabular} &
    \begin{tabular}{@{}l@{}}arbitrary \\ AF\end{tabular} &
    \begin{tabular}{@{}l@{}}$x$-space \\ sparsity \end{tabular} &
    \begin{tabular}{@{}l@{}}$f$-space \\ sparsity \end{tabular} &
    \begin{tabular}{@{}l@{}}$h$-space \\ sparsity \end{tabular} &
    \begin{tabular}{@{}l@{}}blackbox \\ evidence\end{tabular}
    \\%
    \midrule%
    Hallucination \cite{azimi2010batch} & \ding{53} & \ding{51} & \ding{53} & \ding{51} & \ding{51} & \ding{53} & \ding{53} & \ding{53} \\
    LP \cite{gonzalez2016batch} & \ding{51} & \ding{53} & \ding{53} & \ding{51} & \ding{51} & \ding{53} & \ding{53} & \ding{53} \\
    TS \cite{hernandez2017parallel} & \ding{51} & \ding{51} & \ding{51} & \ding{53} & \ding{53} & \ding{53} & \ding{53} & \ding{53} \\
    Decoupled TS \cite{wilson2020efficiently} & \ding{51} & \ding{51} & \ding{51} & \ding{53} & \ding{53} & \ding{51} &\ding{53} & \ding{53} \\
    DPP
    \cite{kathuria2016batched} & \ding{53} & \ding{51} & \ding{53} & \ding{53} & \ding{51} & \ding{53} &\ding{53} & \ding{53} \\
    DPP-TS \cite{nava2022diversified} & \ding{51} & \ding{51} & \ding{51} & \ding{53} & \ding{51} & \ding{53} &\ding{53} & \ding{53} \\
    MC-SAA \cite{balandat2020botorch} & \ding{51} & \ding{51} & \ding{53} & \ding{53} & \ding{51} & \ding{53} &\ding{53} & \ding{53} \\
    GIBBON \cite{moss2021gibbon} & \ding{53} & \ding{51} & \ding{53} & \ding{53} & \ding{51} & \ding{53} &\ding{53} & \ding{53} \\
    TurBO \cite{eriksson2019scalable} & \ding{51} & \ding{53} & \ding{53} & \ding{51} & \ding{51} & \ding{53} &\ding{53} & \ding{53} \\
    SOBER (Ours) & \ding{51} & \ding{51} & \ding{51} & \ding{51} & \ding{51} & \ding{51} & \ding{51} & \ding{51} \\
    \bottomrule
  \end{tabular}
  \end{center}
  \normalsize
  \vspace{-4mm}
\end{table}

\textbf{Challenges in Batch Bayesian Optimisation Tasks.}
Despite its many successes, batch \gls{bo} has several challenges. 
Firstly, \emph{batch size scalability}: The extensive overhead of many batch \gls{bo} methods limits the batch size to be around 10. Scaling to a larger batch size is preferable for a variety of real-world problems. For instance, high-throughput drug discovery might evaluate 384 compounds in a batch experiment \cite{carpentier2016hepatic}. Further, \textit{in-Silico} materials discovery can query thousands of simulations in parallel via computer clusters. As a second challenge, many batch \gls{bo} methods are targeted at \emph{continuous inputs}, yet the examples mentioned above of drug discovery, namely molecules, are inherently discrete. Selecting batch acquisition samples in a discrete space leads to combinatorial explosion with increasing batch size. Lastly, \emph{batch diversity}: scalable methods, such as \gls{ts}  \cite{hernandez2017parallel, kandasamy2018parallelised}, are too exploitative. The two leftmost columns in Figure~\ref{fig:demo} exemplifies typical behaviour -- getting stuck in a local minimum and wasting batch samples in the majority of \gls{ts} methods. This tendency amplifies in noisy and multimodal cases. The larger the batch size we query, the larger the regret becomes, as the batch samples could have been used for exploring other regions.\looseness=-1

\begin{wrapfigure}[20]{R}{0.35\textwidth}
\vspace{-2mm}
\includegraphics[width=0.35\textwidth]{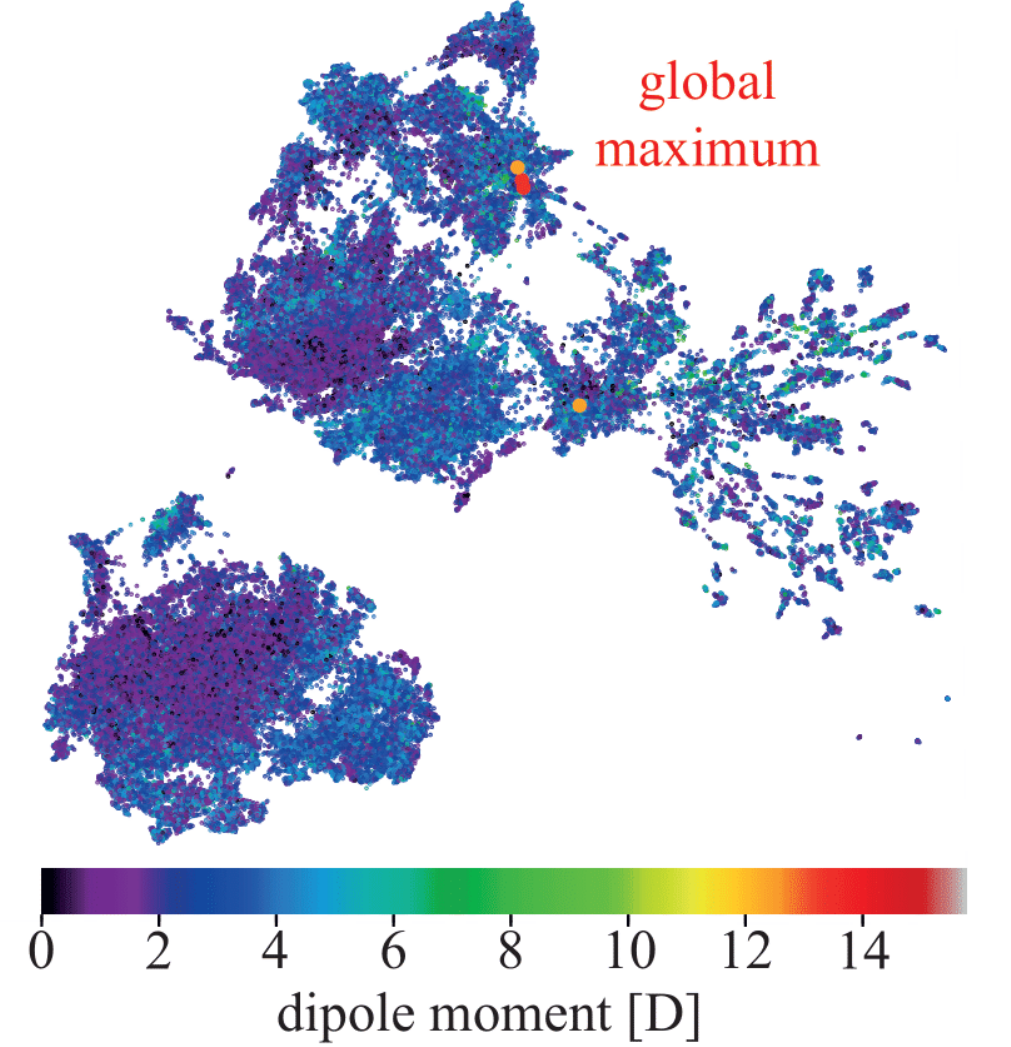}
\caption{Drug discovery is finding the needle in the hey stack. 2D UMAP \cite{mcinnes2018umap} shows only 5 out of 133,055 molecules on the QM9 dataset met desideratum (> 10D).} 
\label{fig:qm9}
\end{wrapfigure}
Figure \ref{fig:qm9} shows drug discovery is a case where the input space is noisy and multimodal. This needle-in-the-haystack situation is challenging for finding the bias-variance trade-off via optimising \gls{gp} hyperparameters. It leads to having two modes: low-noise and high-noise in the hyperposterior space. A low-noise mode regards every tiny change as a peak, whereas high-noise sees every peak as noise. While the low-noise mode tends to get stuck in local minima, the high-noise mode can not find the best drug. Such problems are usually solved by adopting \gls{fbgp}, which uses an ensemble of \gls{gp} models with hyperparameters sampled from hyperposteriors \cite{riis2022bayesian}. Yet, \gls{fbgp} causes significant overhead that challenges for expensive batch \glspl{bo} (e.g. \gls{dpp} \cite{kathuria2016batched}).

\textbf{Challenges in Batch Bayesian Quadrature Tasks.}
Unlike \gls{bo}, \gls{bq} endures `over-exploration' when applied for Bayesian inference. The exploration space in \gls{bq} is defined as the prior distribution, whereas the target distribution to approximate, posterior, is much sharper than prior and shows negligibly small values almost everywhere. Guiding \gls{bq} to explore the posterior mode is key to sample-efficiency.\looseness=-1

\textbf{Contribution.}
We propose the hallucination-free scalable batching: \gls{sober}.
\gls{sober} bridges \gls{bo} and \gls{bq}, combining both merits to solve the above issues of each method --- While exploitative \gls{bo} alleviates over-exploration for \gls{bq}, explorative \gls{bq} mitigates the stuck-in-local-minima issue via diversified sampling for \gls{bo} (see two rightmost columns of Figure \ref{fig:demo}). 
Table \ref{tab:comparison} summarises a comparison with the batch \gls{bo} baselines.
With the given features, we particularly focus on drug discovery (large-scale discrete batch BO with graph/string input, batch \gls{bo} task), and simulation-based inference (simultaneous estimation of both posterior and evidence, batch \gls{bq} task).
Empirically, \gls{sober} shows better sample-efficiency as well as faster wall-clock time computation than 11 baselines in both batch \gls{bo} and \gls{bq} over 12 tasks. We emphasise that \gls{sober} is an extension of both \gls{bo} and \gls{bq} and is compatible with existing methods.

\section{Related Work and Challenges}\label{sec:related}
\textbf{Batch Bayesian Optimisation.}
Batch \glspl{bo} are summarised in Table \ref{tab:comparison} (see Supp. A.1 for primer). While \gls{bo} is an inherently sequential algorithm, assuming queries to an oracle are performed one after another, batch \gls{bo} aims to query multiple locations in one go. However, most \glspl{af} are designed only for querying one point. Classic methods, like hallucination \cite{azimi2010batch} and \gls{lp} \cite{gonzalez2016batch}, tackled this by simulating sequential processes, both of which are successful in small batch size $n$, but not scalable due to large computational overhead. In recent work, MC-SAA \cite{balandat2020botorch} proposed \gls{qmc}-based \gls{af} approximation, and GIBBON \cite{moss2021gibbon} proposed diversified sampling methods using a specific \gls{af}. However, discrete and mixed spaces also present challenges for the above algorithms. The simplest way to maximise \glspl{af} over a discrete space is to take the argmax of all possible candidates. Yet, the higher the dimension and larger the number of categorical classes, the more infeasibly large the combination becomes  (combinatorial explosion). This is particularly challenging for MC-SAA and GIBBON, as both optimise the batch querying points as optimisation variables. Namely, querying large batch sizes requires enumerating all possible permutations of \textit{both batch samples and discrete variables}, leading to a combinatorial explosion. \gls{dpp} \cite{kathuria2016batched} also proposed diversified sampling with rejection sampling, however, it produces prohibitive overhead and is not scalable. Alternatively, there exist \gls{bo} works for discrete and mixed spaces that propose bespoke \gls{af} \cite{baptista2018bayesian, ru2020bayesian, daulton2022bayesian} or special kernel \cite{wan2021think, oh2019combinatorial, deshwal2021mercer, thebelt2022tree}. All consider sequential setting --- we do not compare against sequential \gls{bo}. Rather, these special \gls{af}/kernels are compatible with our method, \gls{sober}.\looseness=-1

As such, the existing scalable discrete batch \gls{bo} methods are \gls{ts}-based (\gls{ts} \cite{hernandez2017parallel}, decoupled \gls{ts} \cite{wilson2020efficiently}, \gls{dpp}-\gls{ts} \cite{nava2022diversified}, TurBO \cite{eriksson2019scalable}). \gls{ts} is approximated by the sequential argmax of random samples over input space, which is completely gradient-free, allowing for scalable batching over a discrete space. Yet, batch \gls{ts} is not so diversified. Decoupled \gls{ts} diversifies \gls{ts} batch samples by decoupled sampling that sparsifies the \gls{gp} function space. Although decoupled sampling yields scalable batch sampling, diversification is not enough as shown in Figure \ref{fig:demo}. \gls{dpp}-\gls{ts} tried to take the best of both worlds of \gls{dpp} and \gls{ts}, and is faster than \gls{dpp}, but still slower than others. TurBO \cite{eriksson2019scalable} introduced multiple local \glspl{bo} bounded with trust regions, and allocates batching budgets based on \gls{ts}. This succeeded in scalable batching via maintaining local \glspl{bo} that are compact, via shrinking trust regions, based on heuristics with many hyperparameters. Selecting hyperpameters is non-trivial and TurBO cannot apply to discrete and non-Euclidean space, for which kernels do not have lengthscale hyperparameters for the trust region update heuristic (e.g. Tanimoto kernel for drug discovery \cite{ralaivola2005graph}). (See Supp. B).\looseness=-1

\textbf{Batch Bayesian Quadrature.}
A primer on \gls{bq} can be found in \cite{hennig2022probabilistic} and Supp. A.2. \gls{bq} shares analogous challenges on batching with \gls{bo}. While \gls{bo} needs accurate approximation around the global optimum, integration approximation error in \gls{bq} is affected by whole function space. Thus, \gls{bq} adopts uncertainty sampling \gls{af} \cite{osborne2012active, gunter2014sampling}. BatchWSABI \cite{wagstaff2018batch} was the first to extend \gls{bq} to a batch setting using \gls{lp} \cite{gonzalez2016batch}. Later, \gls{basq} achieved efficient batching with kernel recombination and is the only method for scalable batching \cite{adachi2022fast}. However, \gls{basq} is over-explorative; log\gls{basq} \cite{adachi2022bayesian} mitigates this by using the log-warped \gls{gp}.

\subsection{Bayesian Alternately Subsampled Quadrature (BASQ)}\label{sec:basq}
The difficulty of scalable batching is the sequential \gls{af} maximisation strategy, of which the large overhead and the continuity assumption hinder scalable discrete space optimisation. The \gls{ts} strategy is too exploitative and less diversified. \gls{basq} \cite{adachi2022fast} achieved scalable batch \gls{bq} with a different approach from existing works. Instead of \gls{af} maximisation, \gls{basq} reframed batch uncertainty sampling (batch \gls{bq}) as \gls{kq}. \gls{kq} is the problem of sample-efficient integration approximation as a weighted sum $\int f(x) \text{d} \pi(x) \approx \textbf{w}_\text{batch}^\top f(\textbf{X}_\text{batch})$, of which the integrand function $f$ belongs to \gls{rkhs}, typically applied to the positive semi-definite kernel. Let the kernel $K$ be the posterior predictive covariance of the \gls{gp}. Surprisingly, minimising worst-case integration error via selecting $n$ weighted samples $(\textbf{w}_\text{batch}, \textbf{X}_\text{batch})$ is \textit{equivalent} to minimising the total predictive uncertainty (proof in \cite{huszar2012optimally}). Thus, solving \gls{kq} can obtain $n$ batch samples which are not only diversified, but also of provably small total \gls{gp} predictive variance (uncertainty sampling). This duality enables us to import the advanced methods from \gls{kq} community. Particularly, \gls{skq} offers multiple advantages over existing batch \gls{bq}/\gls{bo} methods. First, \emph{scalability in batch size}, of which computational complexity is smaller than baselines (see Supp. B). Second, a \textit{gradient-free solver}, which is ideal for non-Euclidean discrete spaces. Third, \textit{no limitation in prior and kernel modelling}, unlike baselines \cite{wagstaff2018batch}. \gls{skq} reformulates the $n$ point selection problem as extracting the subset of discrete probability measure $\pi_\text{batch} = (\textbf{w}_\text{batch}, \textbf{X}_\text{batch})$ from large samples approximating the prior (we refer to empirical measure) $\pi_\text{emp} = (\textbf{w}_\text{rec}, \textbf{X}_\text{rec})$, where $\textbf{X}_\text{batch} \in \mathbb{R}^{n \times d}$, $\textbf{X}_\text{rec} \in \mathbb{R}^{N \times d}$, $\textbf{X}_\text{batch} \subset \textbf{X}_\text{rec}$, $N \gg n$. Empirical measure is constructed by sampling from $\pi$, $\textbf{X}_\text{rec} \sim \pi$, and its expectation approximates the mean of $\pi$ well, $\int f \text{d} \pi \approx \textbf{w}_\text{rec}^\top f(\textbf{X}_\text{rec}) \approx \textbf{w}_\text{batch}^\top f(\textbf{X}_\text{batch})$ for $f$ in the \gls{rkhs}. This formulation can be interpreted as minimising \gls{mmd} \cite{gretton2012kernel} between $\pi_\text{emp}$ and $\pi_\text{batch}$ \cite{huszar2012optimally}. The \gls{mmd} distance is widely recognised for measuring the distance between distributions. 
As such, \gls{skq} is a sparse discretisation problem of an uncertain region subject to approximate the original $\pi$ with given $n$ samples with weights.\looseness=-1

Amongst \glspl{skq}, \gls{rchq} \cite{hayakawa2021positively} achieved provably state-of-the-art convergence rate of integral approximation as well as computationally-tractable complexity. This rate and complexity come from two approaches: Nyström method \cite{drineas2005on} and kernel recombination \cite{tch15}. The Nyström method approximates the kernel using randomised \gls{svd} \cite{halko2011finding} of the Gram matrix. \gls{svd} eigendecomposes the Gram matrix $K(\textbf{X}_\text{rec}, \textbf{X}_\text{rec}) = \textbf{U} \text{diag}(\boldsymbol\Lambda) \textbf{U}^\top$, then the test functions, defined as $\varphi_j(x) := u_j^\top K(\textbf{X}_\text{nys}, x)$, where $u_j \in \textbf{U}$ is an eigenvector, $\lambda_j \in \text{diag}(\boldsymbol\Lambda)$ is an eigenvalue, and $\textbf{X}_\text{nys} \in \mathbb{R}^{M \times d}$ drawn from $\pi$, can approximate the kernel $K(x, x^\prime) \approx \sum_{j=1}^{n-1} \lambda_j^{-1} \varphi_j(x) \varphi_j(x^\prime)$. \gls{rchq} constructs sparse \gls{kq} rules via $n - 1$ test functions, permitting incorporation of the spectral decay in the kernel for faster convergence. \textit{Kernel recombination} is the algorithm to solve such \gls{skq} problems with best known complexity. This algorithm is completely gradient-free. In summary, the advantages of \gls{basq} derive from: (1) Scalability, (2) gradient-free solver for discrete and non-Euclidean spaces, and (3) no limitation in prior and kernel selection, as comes from reformulation as \gls{skq}. Constructing $\pi_\text{emp}$ is just sampling, so any kind of distribution can be used. The Nyström approximation only requires Gram matrix, which any kernel can compute.

\section{Proposed Method: SOBER} \label{sec:sober}

\subsection{Global Optimisation as Bayesian Quadrature: Duality in Probability Measure}
\begin{figure}
    \includegraphics[width=1\textwidth]{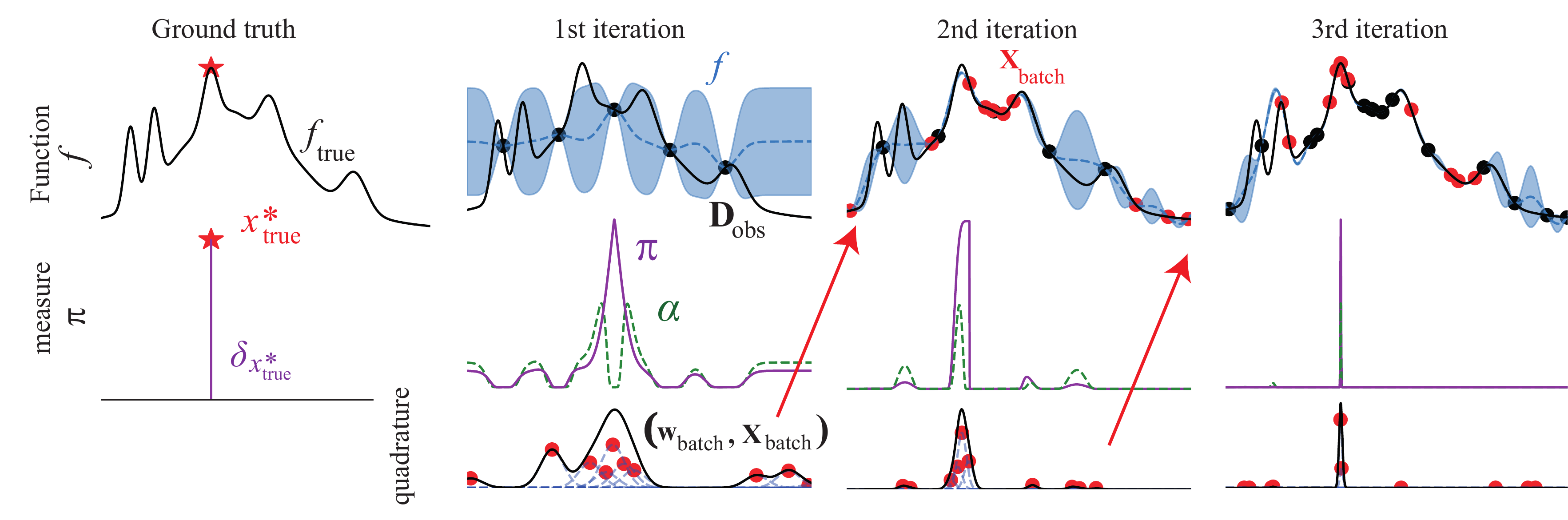}
    \caption{SOBER algorithm. Finding the location of global maximum $x^*_\text{true}$ is equivalent to finding the delta distribution $\delta_{x^*_\text{true}}$. Based on the surrogate $f$, we approximate the probability of global maximum $P(x^*_\text{true} \mid \textbf{D}_\text{obs})$ as $\pi$ defined in \S\ref{sec:pi}. We can also set the user-defined acquisition function $\alpha$ to adjust batch samples (max-value entropy search in this case).
    Kernel quadrature gives a weighted point set $(\textbf{w}_\text{batch}, \textbf{X}_\text{batch})$ that makes a discrete probability measure approximating $\pi$ in a way that is adaptive to the current \gls{gp} \cite{huszar2012optimally}.
    Here, we have used a weighted kernel density estimation based on $(\textbf{w}_\text{batch}, \textbf{X}_\text{batch})$ to approximately visualise the kernel quadrature.
    Over iterations, $\pi$ shrinks toward global maximum, which ideally becomes the delta function in a single global maximum case.
    }
    \label{fig:sober}
\end{figure}

We reframe batch \gls{bo} as a batch \gls{bq} problem. Consider the following dual formulation \cite{rudi2020finding}:
\begin{equation}
    x^*_\text{true} = \arg \max_x f_\text{true}(x) \quad \xLeftrightarrow{\text{dual}} \quad
    \delta_{x^*_\text{true}} \in \mathop\mathrm{arg\,max}_{\pi} \int f_\text{true}(x) \text{d}\pi(x),  \label{eq:dual}
\end{equation}
where $\delta_x$ is the delta distribution at $x$, $\pi$ is a probability distributions over the $x^*_\text{true}$, and $x^*_\text{true}$ is the location of the global maximum of $f_\text{true}$. Figure \ref{fig:sober} illustrates the algorithm flow.

\textbf{Why necessary?}
This formulation is defined as black-box integration. We can then solve the batch \gls{bo} problem as batch \gls{bq} using \gls{basq} to harvest the benefits: (1) scalable diversified batching, (2) applicability to discrete and non-Euclidean space, and (3) flexibility of \gls{gp} models with arbitrary kernel and priors over input. These benefits are unachievable with existing methods as previously mentioned.\looseness=-1

\textbf{How is it different from BASQ?}
Eq. \eqref{eq:dual} updates $\pi$ over each iteration, whereas \gls{basq} keeps $\pi$ unchanged. The more accurately the \gls{gp} surrogate approximates $f_\text{true}$ through batch-sequential updates, the ``narrower'' $\pi$ gets. Ideally, $\pi$ eventually reaches the global maximum $\delta_{x^*_\text{true}}$ in a single global maximum case.
As such, solving Eq. \eqref{eq:dual} by updating $\pi$ is equivalent to finding the maximum $x^*_\text{true}$. 
While \gls{basq} is a pure exploration algorithm, \gls{sober} introduces exploitation by shrinking $\pi$.

\subsection{Batch Selection as Kernel Quadrature} \label{sec:batch}
The batch selection in \gls{sober} is performed by \gls{rchq}. We propose the \textit{objective-}\gls{rchq}, solving:
\begin{subequations}
\begin{align}
    &\text{Find an $n$-point subset} \quad 
    \textbf{X}_\text{batch} \subset \textbf{X}_\text{rec},\\
    \text{s.t.} \quad &\text{max} \quad \textbf{w}_\text{batch}^\top \alpha(\textbf{X}_\text{batch}), \label{eq:obj}\\ 
    &\textbf{w}_\text{batch}^\top \textbf{1} = \textbf{w}_\text{rec}^\top \textbf{1} = 1, \quad \textbf{w}_\text{batch} \ge \textbf{0}, \quad \textbf{w}_\text{batch}^\top \varphi_i(\textbf{X}_\text{batch})=
    \textbf{w}_\text{rec}^\top \varphi_i(\textbf{X}_\text{rec}), \quad 1\le i<n, \label{eq:rec}
\end{align}
\end{subequations}
where $\alpha$ is an arbitrary \gls{af}. As with the known constraints of $n - 1$ \gls{kq} rules with Nyström approximation in Eq. \eqref{eq:rec}, we added the \gls{af} maximisation term Eq. \eqref{eq:obj}. 
There exist theoretical guarantees for specific choices of \gls{af}s depending on the spectral decay of an integral operator determined by the pair $(K, \pi)$ \cite{adachi2022fast, hayakawa2023sampling}.
Empirically, the convergence as $n$ increase is fast even without the \gls{af} term.
We can utilise this degree of freedom for incorporating \gls{af} tailored for specific domain.

\subsection{Sequentially Updating \texorpdfstring{$\pi$}{p}} \label{sec:pi}
$\pi$ is vital in defining the feasible region by exploiting the information from the surrogate model $f$. We wish to design $\pi$ to shrink towards the global maximum, which ideally becomes the delta function in a single global maximum case.
We discuss two variants; \gls{ts} and \gls{lfi}.

\textbf{TS-based.}
\gls{sober}-\gls{ts} adopts the maximum of the current \gls{gp} surrogate $x^*$, $P(x^*|\textbf{D})$ as $\pi$. This definition is well known as \gls{ts}. We prepare $N$ candidates with parallel decoupled \gls{ts} \cite{wilson2020efficiently}, which alleviates the sampling overhead and fosters batch diversity via sparsifying the sampled functions from \gls{gp} posterior ($f$-space sparsity). As such, \gls{sober}-\gls{ts} can be understood as re-selecting the \gls{ts} samples that can maximise the sum of the \gls{af} and minimise the predictive variance. However, \gls{ts} cannot provide the closed-form $P(x^*|\textbf{D})$ distribution for faster sampling. 

\textbf{LFI-based.}
Considering $\pi$ definition under uncertain global maximiser $x^*$, $x^*_\text{true}$ might be any location with values possibly larger than the current maximum $\eta := \max \mathbb{E} [f(x^*)]$. Given this definition, we can define the ``tentative'' likelihood over input space $P(f(x) \geq \eta|x, \textbf{D})$ \cite{gutmann2016bayesian}, given by:
\begin{align}
L(x|\theta, \sigma^2_n, \eta) := P \big( f(x|\textbf{D} \big) \geq \eta) \propto \Phi \left[\frac{m(x|\theta) - \eta}{\sqrt{C(x,x|\theta)}} \right], \label{eq:templik}
\end{align}
where the $\Phi$ is \gls{cdf} of the standard normal distribution. This likelihood definition is the same as the \gls{pi} \gls{af} \cite{kushner1964new}. 
Ideally, $\pi$ can eventually reach the global maximum $\delta_{x^*_\text{true}}$ in a single global maximum case, where $\eta = f_\text{true}(x^*_\text{true}) = \mathbb{E}[f(x^*_\text{true})]$.
Now we can estimate posterior belief of the maximum location $\pi$ via the following Bayes’ rule, $\pi(x) \propto L(x) \pi^\prime(x)$. For each iteration, $L(\cdot|\theta, \sigma^2_n, \eta)$ is updated and $\pi^\prime$ is $\pi$ of the previous iteration. In the beginning, $\pi^\prime$ is the prior belief. In the typical \gls{bo} setting, we use the bounded uniform prior, but we can incorporate the experts' knowledge via a stronger prior. We do not examine the sensitivity to prior $\pi^\prime$ in this paper, but the following papers show strong empirical performance of a stronger prior \cite{mayr2022learning, hvarfner2022pi}.
Closed-form expression of \gls{sober}-\gls{lfi} can offer faster sampling.

\textbf{How to sample from \texorpdfstring{$\pi$}{p}.}
\gls{sober} handles discrete, continuous, or mixed inputs. The difference is the sampler for empirical measure $\pi_\text{emp}$.
The simplest scenario is if all discrete candidates are available \textit{a priori} and enumerable. As \gls{rchq} accepts weighted samples $\pi_\text{emp} =(\textbf{w}_\text{rec}, \textbf{X}_\text{rec})$ as importance sampling, sampling is just computing the weights $\textbf{w}_\text{rec} = \pi(\textbf{X}_\text{rec}) / \big[ \pi(\textbf{X}_\text{rec})^\top \textbf{1} \big]$. If all combinations are innumerable or unavailable, we sample $\textbf{X}_\text{rec}$ from the discrete prior $\pi^\prime$, which is user-defined. Once sampled, the procedure is the same: compute $\textbf{w}_\text{rec}$, then pass the empirical measure $\pi_\text{emp}$ to \gls{rchq}. We update the hyperparameters of the prior $\pi^\prime$ via \gls{mle} from the weighted sample $(\textbf{w}_\text{rec}, \textbf{X}_\text{rec})$.
A continuous space can be regarded as an innumerable discrete space, so it can be handled similarly. The only difference is the prior update. We use weighted \gls{kde} for the update, for speed and flexibility.
Mixed space is the combination of discrete and continuous space, so the prior is the combination of both. Importantly, the prior does not need to precisely approximate $\pi$ as the importance weights $\textbf{w}_\text{rec}$ will correct the difference. See Supp. D for details.\looseness=-1

\subsection{Auxiliary Algorithms} \label{sec:aux}
The fundamental idea of \gls{sober} is simple: reframe batch \gls{bo} as a batch \gls{bq} problem. While objective-\gls{rchq} plays a role of exploration, shrinking $\pi$ contributes to exploitation. The balance between exploration and exploitation can be adjustable via \gls{af} in objective-\gls{rchq}, $\pi$ definitions, or kernel selection, similarly to standard \gls{bo}. While the above functionalities suffice minimal components of \gls{sober}, the following auxiliary ones help bringing out the full potential of \gls{sober} on each case.

\textbf{Noisy Functions.} \label{sec:fast maginalisation}
As seen in Figure \ref{fig:qm9}, noisy functions need \gls{fbgp} modelling \cite{riis2022bayesian} but is expensive. We propose \textbf{quadrature distillation} to mitigate this. As \gls{fbgp} is the \gls{mc} integration for marginal predictive posterior over hyperposterior (blackbox function), we can reframe this as another \gls{bq} problem. This reformulation extracts a small subset of weighted hypersamples from random ones. Such a distilled subset can offer sample-efficient integral approximattion over the hyperposterior, which can accelerate various expensive integrations, such as fast \gls{fbgp}, fast information-theoretic \gls{af}, and fully Bayesian \gls{lfi} formulation (see Supp. C).

\textbf{Non-Smooth Functions.}
\gls{rchq} exploits spectral decay of kernel via Nyström method for faster convergence. However, non-smooth or sharp functions have a long-tailed decay in eigenvalues, worsening the convergence rate of quadrature algorithms. We propose \textbf{automatic KQ selection} algorithm to cope with this. \gls{skq} is not only \gls{rchq} — kernel thinning \cite{dwivedi2021kernel}is also subsample-based and independent on the spectral decay. Which performs better depends on the level of smoothness. Importantly, the quadrature performance is quantifiable as worst-case integration error, thus we can automate the selection of two \gls{skq} algorithms by monitoring the error (see Supp. D).

\textbf{Simulation-based Inference.}
As \gls{lfi} is originally introduced to solve simulation-based inference, now our \gls{sober}-\gls{lfi} is also capable of solving simulation-based inference. Sampling from $\pi$ efficiently squeezes the region to be explored toward only the vicinity of posterior mode. As such, \gls{bo} reformulation can introduce exploitation to \gls{bq} via $\pi$ update. When compared to the original \gls{bo}-based \gls{lfi} (BOLFI) \cite{gutmann2016bayesian}, \gls{sober} has two benefits; evidence estimation and exact posterior estimation. While BOLFI is designed to approximate only the posterior distribution using the approximated likelihood definition, \gls{sober} can estimate both the posterior and model evidence in one go, using the exact likelihood definition based on \gls{basq}. We propose \textbf{dual GPs}; one for sampling with BOLFI, one for \gls{bq} modelling with \gls{basq}. (see Supp. E)

\subsection{Summary of Contribution}
We reformulated the batch \gls{bo} task as the dual problem defined by Eq.\ \eqref{eq:dual}. 
Now, estimating the global maximum is equivalent to updating $\pi$. We introduced two variants: \gls{ts} and \gls{lfi}. Both offer approximation of $\pi$ using the information of the current surrogate model $f$. In either case, once the empirical measure $\pi_\text{emp} = (\textbf{w}_\text{rec}, \textbf{X}_\text{rec})$ is constructed by sampling from $\pi$, the `objective-\gls{rchq}' chooses batch samples that minimise the \gls{gp} posterior variance and maximise the user-defined \gls{af}. Additionally, quadrature distillation provides fast \gls{fbgp} for noisy functions, automatic \gls{kq} selection permits the best of both worlds over \gls{rchq} and kernel thinning for non-smooth or sharp functions, and dual \glspl{gp} offer simulation-based inference. Note that the auxiliary algorithms are additional augmentations of general \gls{gp}-based \gls{bo} against the shared challenges. These design choices are simple, whether or not the objective function is noisy, non-smooth, or sharp. If the additional overhead allows, we can use all in default, but \gls{sober} works even without these (see Supp. D).

\section{Link between Global Optimisation and Quadrature}
\subsection{Theoretical Background}
Similar approaches to ours can be seen in existing batch \glspl{bo}, such as TurBO \cite{eriksson2019scalable} (shrinking trust regions), batch \gls{ts} \cite{kandasamy2018parallelised}, and \gls{dpp}-\gls{bo} \cite{kathuria2016batched}. For instance, the motivation of \gls{dpp} is to diversify the batch samples \cite{nava2022diversified}, but the underlying idea can be linked with quadrature. As explained in Sec.~\ref{sec:basq}, quadrature can be seen as sparsely discretising an uncertain region, which again can be regarded as pure exploration with diversified batch samples for \gls{bo} via Eq. \eqref{eq:dual}. For a more detailed theoretical background, we can start with classical batch \gls{ts}. First of all, \cite{kandasamy2018parallelised} show that the batch \gls{ts} (synTS in their paper) satisfies the reasonable order of \gls{br} (their Theorem 2). Their proof is essentially given by estimating $\frac1t \sum_{i=1}^t \frac1N \sum_{j=1}^N \mathbb{E}[f_\text{true}(x^*_\text{true}) - f_\text{true}(x_{i,j})]$, where $(x_{i,j}) \in \textbf{X}_{\text{batch}, i}$ is the batch sample at the $i$-th iteration. We can view the value $f_\text{true}(x^*_\text{true}) - \frac1N\sum_{j=1}^N f_\text{true}(x_{i,j})$ as the difference between $f_\text{true}(x^*_\text{true})$ and the \gls{mc} estimate of the integral of $f_\text{true}$ over $\pi_i$.
In \gls{sober}, we approximate this integral by an $n$-point weighted subset of $\textbf{X}_{\text{batch}, i}$ at each iteration using kernel recombination — on which we can prove the integral error (see Theorem 1 in \cite{adachi2022fast})\footnote{Although, more investigation is needed to obtain end-to-end convergence guarantees.}. Namely, we can quantify how close our integral estimate $\textbf{w}_\text{batch}^\top f(\textbf{X}_\text{batch})$ is to the true integral $\int f_\text{true}(x) \text{d} \pi(x)$ or the empirical mean $\textbf{w}_\text{rec}^\top f(\textbf{X}_\text{rec})$. 
Similar discussion can be done with \gls{dpp}-\gls{bo} \cite{kathuria2016batched, nava2022diversified} for general \glspl{af}.

\subsection{Empirical Analysis}
\begin{table}
    \caption{Experimental Setup. \gls{bo}/\gls{bq} refers to which task is to be solved, syn./real refers to the problem whether synthetic or real-world ones. The dimensions represent the number of dimensions over input space categorised into continuous (cont.), categorical (cat.), and binary (bin.), total refers to the summation of each dimension. Batch is the batch size $n$, and prior is $\pi^\prime$. For prior, Bernoul. and Categor. refer to Bernoulli and categorical distributions with equal weights. (see Suppl. F)
    }
    \small
    \label{tab:exp}
    \centering
    \begin{center}
    \begin{tabular}{p{14mm}p{4mm}p{4mm}p{18mm}p{3mm}p{2mm}p{4mm}p{4mm}p{5mm}p{10mm}p{12mm}p{10mm}}
    \toprule
    & & & & \multicolumn{4}{c}{dimensions $d$}
    & & & \multicolumn{2}{c}{prior $\pi^\prime$} \\
    \cmidrule(lr){5-8} \cmidrule(lr){11-12}
    experiments &
    \begin{tabular}[c]{@{}l@{}}BO\\ /BQ\end{tabular} & \begin{tabular}[c]{@{}l@{}}syn.\\ /real\end{tabular} & objective & cont. & cat. & bin. & total & 
    \begin{tabular}[c]{@{}c@{}}batch\\ $n$\end{tabular} &
    \begin{tabular}[c]{@{}c@{}}kernel\\ $K$\end{tabular} & cont. & disc. \\
    \midrule
    Ackley & \multirow{11}{*}{BO} & \multirow{4}{*}{syn} & $\log_{10} \min(f)$ & 3 & - & 20 & 23 & 200 & RBF & $\mathcal{U}(-1,1)$ & Bernoul. \\
    Rosenbrock &  & & $\log_{10} \min(f)$
    & 1 & 6 & - & 7 & 100 & RBF & $\mathcal{U}(-4,11)$ & Categor. \\
    Hartmann &
    &                      & $\log_{10} \max(f)$ & 6 & - & - & 6 & 100 & RBF & $\mathcal{U}(0,1)$ & - \\
    Snekel      &                      &                      & $\log_{10} \max(f)$ & 4 & - & - & 4 & 100 & RBF & $\mathcal{U}(0,10)$ & - \\
    \cmidrule(lr){3-12}
    Pest       &                      & \multirow{6}{*}{real} & $\min(f)$ & - & 15 & - & 15 & 200 & RBF & - & Categor. \\
    MaxSat      &                      & & $\min(f)$                     & - & - & 28 & 28 & 200 & RBF & - & Bernoul. \\
    Ising       &                      & & $\log_{10} \min(f)$                     & - & - & 24 & 24 & 100 & RBF & - & Bernoul. \\
    SVM         &                      & & $\min(f)$                     & 3 & - & 20 & 23 & 200 & RBF & $\mathcal{U}(0,1)$ & Bernoul. \\
    Malaria     &                      & & $ \log_{10} \min(f)$                     & - & - & 2048 & 2048 & 100 & Tanimoto & - & Categor.\\
    Solvent     &                      & & $- \max\log_{10}f$                     & - & - & 2048 & 2048 & 200 & Tanimoto & - & Categor. \\
    \cmidrule(lr){2-12}
    2 RC        &  \multirow{2}{*}{BQ} & \multirow{2}{*}{real} & $\int f(x) \text{d} \pi^\prime(x)$ & 6 & - & - & 6 & 100 & RBF & Gaussian & - \\
    5 RC        &                      &  & $\int f(x) \text{d} \pi^\prime(x)$     & 12 & - & - & 12 & 100 & RBF & Gaussian & -\\
    \bottomrule
    \end{tabular}
    \end{center}
    \normalsize
    \vspace{-4mm}
\end{table}
We empirically investigate the relationship between regret and $\pi$-shrinkage. We consider the following two metrics for $\pi$-shrinkage; \gls{mv} $\mathbb{V}\text{ar}[\pi(x)]$ and \gls{md} $|x^*_\text{true} - \mathbb{E}[\pi(x)]|$. \gls{mv} corresponds to the $\pi$-shrinkage, of which smaller value indicates shrinking. \gls{md} represents the Euclidean distance between the mean of $\pi$ and the true global maximum $x^*_\text{true}$.
We compared these two metrics against \gls{br}, $\text{BR} := | y^*_\text{true} - \textbf{w}_\text{batch}^\top f_\text{true}(\textbf{X}_\text{batch}) |$. \gls{br} is the batch estimation regret (referred as \gls{br} for Theorem 2 in \cite{kandasamy2018parallelised}).
Experimental results (see Supp. F.) show the strong linear correlations between all of these 3 metrics (> 0.95), clearly explaining the $\pi$-shrinkage is a good measure of \gls{br} as the dual objective in Eq. \eqref{eq:dual}. 
In other words, $\pi$ (\gls{mc} estimate of $x^*$) shrinks toward true global maximum $x^*_\text{true}$ with being smaller variance (more confident), and both linearly correlated to minimising \gls{br}, can be visually confirmed in Figure \ref{fig:sober}.

\section{Experiments}
We evaluate the sample efficiency and overhead of \gls{sober} on \textbf{12 experiments against 11 baselines} (8 baselines for \gls{bo}; random, \gls{ts}, decoupled \gls{ts}, \gls{dpp}-\gls{ts}, TurBO, GIBBON, hallucination, \gls{lp}, and 3 baselines for \gls{bq}; batchWSABI, BASQ, and logBASQ). 12 experimental conditions are summarised in Table \ref{tab:exp} (10 experiments for BO, 4 synthetic and 6 real-world datasets, and 2 real-world experiments for BQ). See Supp. F for full experimental details. Our code is built upon PyTorch-based libraries \cite{paszke2019pytorch, gardner2018gpytorch, balandat2020botorch, griffiths2022gauche}. All experiments were averaged over 10 repeats, computed in parallel with multicore CPUs\footnote{Performed on MacBook Pro 2019, 2.4 GHz 8-Core Intel Core i9, 64 GB 2667 MHz DDR4} for a fair comparison, although GPUs can accelerate \gls{sober}. As explained in Sec.~\ref{sec:related}, TurBO, GIBBON, Hallucination, and \gls{lp} suffer from a combinatorial explosion in discrete space. To enable the comparison, we adopt thresholding\footnote{We optimise discrete variables as continuous ones, then the solutions are classified via nearest neighbours.} only for their discrete or mixed experiments. Most algorithms cannot be applied for various reasons to drug discovery tasks (see Table \ref{tab:conv}).

\textbf{Results.}
\begin{table}
    \caption{Experimental BO results on convergence of objectives atfter 15 iterations. Objective values are average $\pm$ 1 standard error over 10 repeated runs. See Table~\ref{tab:exp} for the objectives of each task. Mean rank is the average rank over 10 datasets for each method.}
    \small
    \label{tab:conv}
    \centering
    \begin{center}
    \begin{tabular}{p{12mm}p{7mm}p{7mm}p{7mm}p{7mm}p{7mm}p{7mm}p{7mm}p{7mm}p{7mm}p{7mm}|p{7mm}}
    \toprule
    baselines & 
    Ackley &
    \begin{tabular}[c]{@{}l@{}}Rosen-\\ brock\end{tabular} &
    \begin{tabular}[c]{@{}l@{}}Hart-\\ mann\end{tabular} &
    Shekel &
    Pest &
    MaxSat &
    Ising & SVM & 
    \begin{tabular}[c]{@{}l@{}}Mal-\\ aria\end{tabular} &
    \begin{tabular}[c]{@{}l@{}}Sol-\\ vent\end{tabular} &
    \begin{tabular}[c]{@{}l@{}}Mean\\ rank\end{tabular}
    \\
    \midrule
    Random & 
    \begin{tabular}[c]{@{}l@{}}0.232\\ ± 0.01\end{tabular} &
    \begin{tabular}[c]{@{}l@{}}3.509\\ ± 0.02\end{tabular} &
    \begin{tabular}[c]{@{}l@{}}0.448\\ ± 0.01\end{tabular} &
    \begin{tabular}[c]{@{}l@{}}-1.280\\ ± 0.06\end{tabular} &
    \begin{tabular}[c]{@{}l@{}}8.514\\ ± 0.06\end{tabular} &
    \begin{tabular}[c]{@{}l@{}}-25.50\\ ± 0.30\end{tabular} &
    \begin{tabular}[c]{@{}l@{}}-2.008\\ ± 0.12\end{tabular} &
    \begin{tabular}[c]{@{}l@{}}0.363\\ ± 0.01\end{tabular} &
    \begin{tabular}[c]{@{}l@{}}-2.593\\ ± 0.03\end{tabular} &
    \begin{tabular}[c]{@{}l@{}}-0.972\\ ± 0.01\end{tabular} &
    7.1
    \\
    TS &
    \begin{tabular}[c]{@{}l@{}}0.093\\ ± 0.01\end{tabular} &
    \begin{tabular}[c]{@{}l@{}}1.382\\ ± 0.31\end{tabular} &
    \begin{tabular}[c]{@{}l@{}}0.505\\ ± 0.01\end{tabular} &
    \begin{tabular}[c]{@{}l@{}}-0.195\\ ± 0.01\end{tabular} &
    \begin{tabular}[c]{@{}l@{}}9.397\\ ± 0.12\end{tabular} &
    \begin{tabular}[c]{@{}l@{}}-30.37\\ ± 0.54\end{tabular} &
    \begin{tabular}[c]{@{}l@{}}0.027\\ ± 0.07\end{tabular} &
    \begin{tabular}[c]{@{}l@{}}0.349\\ ± 0.01\end{tabular} &
    \begin{tabular}[c]{@{}l@{}}-2.703\\ ± 0.03\end{tabular} &
    \begin{tabular}[c]{@{}l@{}}-0.990\\ ± 0.01\end{tabular} &
    6.0
    \\
    \begin{tabular}[c]{@{}l@{}}decoupled\\ TS\end{tabular} &
    \begin{tabular}[c]{@{}l@{}}0.054\\ ± 0.01\end{tabular} &
    \begin{tabular}[c]{@{}l@{}}2.490\\ ± 0.14\end{tabular} &
    \begin{tabular}[c]{@{}l@{}}0.502\\ ± 0.01\end{tabular} &
    \begin{tabular}[c]{@{}l@{}}0.072\\ ± 0.05\end{tabular} &
    \begin{tabular}[c]{@{}l@{}}7.911\\ ± 0.01\end{tabular} &
    \begin{tabular}[c]{@{}l@{}}-32.53\\ ± 0.25\end{tabular} &
    \begin{tabular}[c]{@{}l@{}}-0.157\\ ± 0.01\end{tabular} &
    \begin{tabular}[c]{@{}l@{}}0.348\\ ± 0.01\end{tabular} &
    \multicolumn{2}{c|}{
    \begin{tabular}[c]{@{}c@{}}incompatible\\ RFF kernel\end{tabular}
    }&
    5.0
    \\
    DPP-TS &
    \begin{tabular}[c]{@{}l@{}}0.071\\ ± 0.01\end{tabular} &
    \begin{tabular}[c]{@{}l@{}}0.976\\ ± 0.26\end{tabular} &
    \begin{tabular}[c]{@{}l@{}}0.495\\ ± 0.01\end{tabular} &
    \begin{tabular}[c]{@{}l@{}}-0.821 \\ ± 0.08\end{tabular} &
    \begin{tabular}[c]{@{}l@{}}7.963\\ ± 0.05\end{tabular} &
    \begin{tabular}[c]{@{}l@{}}-31.18\\ ± 0.45\end{tabular} &
    \begin{tabular}[c]{@{}l@{}}-0.320\\ ± 0.01\end{tabular} &
    \begin{tabular}[c]{@{}l@{}}0.357\\ ± 0.01\end{tabular} &
    \multicolumn{2}{c|}{
    \begin{tabular}[c]{@{}c@{}}prohibitively\\ slow (>7 days)\end{tabular}
    }&
    6.0
    \\
    TurBO &
    \begin{tabular}[c]{@{}l@{}}0.200\\ ± 0.01\end{tabular} &
    \begin{tabular}[c]{@{}l@{}}\textbf{-0.017}\\ \textbf{± 0.01}\end{tabular} &
    \begin{tabular}[c]{@{}l@{}}0.499\\ ± 0.01\end{tabular} &
    \begin{tabular}[c]{@{}l@{}}0.458\\ ± 0.01\end{tabular} &
    \begin{tabular}[c]{@{}l@{}}9.309\\ ± 0.03\end{tabular} &
    \begin{tabular}[c]{@{}l@{}}-18.25\\ ± 0.47\end{tabular} &
    \begin{tabular}[c]{@{}l@{}}0.169\\ ± 0.02\end{tabular} &
    \begin{tabular}[c]{@{}l@{}}0.362\\ ± 0.01\end{tabular} &
    \multicolumn{2}{c|}{
    \begin{tabular}[c]{@{}c@{}}no lengthscale\\ in kernel\end{tabular}
    } &
    7.0
    \\
    GIBBON &
    \begin{tabular}[c]{@{}l@{}}0.007\\ ± 0.02\end{tabular} &
    \begin{tabular}[c]{@{}l@{}}2.800\\ ± 0.20\end{tabular} &
    \begin{tabular}[c]{@{}l@{}}0.368\\ ± 0.02\end{tabular} &
    \begin{tabular}[c]{@{}l@{}}0.375\\ ± 0.03\end{tabular} &
    \begin{tabular}[c]{@{}l@{}}8.890\\ ± 0.05\end{tabular} &
    \begin{tabular}[c]{@{}l@{}}-33.58\\ ± 1.08\end{tabular} &
    \begin{tabular}[c]{@{}l@{}}-0.228\\ ± 0.01\end{tabular} &
    \begin{tabular}[c]{@{}l@{}}0.365\\ ± 0.01\end{tabular} &
    \multicolumn{2}{c|}{
    \begin{tabular}[c]{@{}c@{}}combinatorial\\ explosion\end{tabular}
    } &
    6.4
    \\
    \begin{tabular}[c]{@{}l@{}}Halluci-\\ nation\end{tabular} &
    \begin{tabular}[c]{@{}l@{}}0.142\\ ± 0.01\end{tabular} &
    \begin{tabular}[c]{@{}l@{}}0.704\\ ± 0.28\end{tabular} &
    \begin{tabular}[c]{@{}l@{}}0.504\\ ± 0.01\end{tabular} &
    \begin{tabular}[c]{@{}l@{}}-0.013\\ ± 0.02\end{tabular} &
    \begin{tabular}[c]{@{}l@{}}8.176\\ ± 0.04\end{tabular} &
    \begin{tabular}[c]{@{}l@{}}-31.05\\ ± 0.46\end{tabular} &
    \begin{tabular}[c]{@{}l@{}}-2.345\\ ± 0.03\end{tabular} &
    \begin{tabular}[c]{@{}l@{}}0.327\\ ± 0.01\end{tabular} &
    \multicolumn{2}{c|}{
    \begin{tabular}[c]{@{}c@{}}prohibitively\\ slow (>7 days)\end{tabular}
    } &
    4.6
    \\
    LP &
    \begin{tabular}[c]{@{}l@{}}-0.063\\ ± 0.01\end{tabular} &
    \begin{tabular}[c]{@{}l@{}}3.727\\ ± 0.19\end{tabular} &
    \begin{tabular}[c]{@{}l@{}}0.511\\ ± 0.01\end{tabular} &
    \begin{tabular}[c]{@{}l@{}}-0.189\\ ± 0.08\end{tabular} &
    \begin{tabular}[c]{@{}l@{}}9.111\\ ± 0.11\end{tabular} &
    \begin{tabular}[c]{@{}l@{}}-20.98\\ ± 1.39\end{tabular} &
    \begin{tabular}[c]{@{}l@{}}-1.527\\ ± 0.28\end{tabular} &
    \begin{tabular}[c]{@{}l@{}}0.343\\ ± 0.01\end{tabular} &
    \multicolumn{2}{c|}{
    \begin{tabular}[c]{@{}c@{}}Non-Euclidean\\ space\end{tabular}
    } &
    5.9
    \\
    \begin{tabular}[c]{@{}l@{}}SOBER\\ -TS\end{tabular} &
    \begin{tabular}[c]{@{}l@{}}-0.008\\ ± 0.02\end{tabular} &
    \begin{tabular}[c]{@{}l@{}}1.934\\ ± 0.250\end{tabular} &
    \begin{tabular}[c]{@{}l@{}}0.513\\ ± 0.01\end{tabular} &
    \begin{tabular}[c]{@{}l@{}}0.242\\ ± 0.02\end{tabular} &
    \begin{tabular}[c]{@{}l@{}}8.377\\ ± 0.01\end{tabular} &
    \begin{tabular}[c]{@{}l@{}}-32.29\\ ± 0.27\end{tabular} &
    \begin{tabular}[c]{@{}l@{}}-2.749\\ ± 0.01\end{tabular} &
    \begin{tabular}[c]{@{}l@{}}0.349\\ ± 0.01\end{tabular} &
    \multicolumn{2}{c|}{
    \begin{tabular}[c]{@{}c@{}}incompatible\\ RFF kernel\end{tabular}
    } &
    3.9
    \\
    \begin{tabular}[c]{@{}l@{}}SOBER\\ -LFI\end{tabular} &
    \begin{tabular}[c]{@{}l@{}}\textbf{-2.180}\\ \textbf{± 0.01}\end{tabular} &
    \begin{tabular}[c]{@{}l@{}}0.653\\ ± 0.17\end{tabular} &
    \begin{tabular}[c]{@{}l@{}}\textbf{0.514}\\ \textbf{± 0.01}\end{tabular} &
    \begin{tabular}[c]{@{}l@{}}\textbf{0.713}\\ \textbf{± 0.01}\end{tabular} &
    \begin{tabular}[c]{@{}l@{}}\textbf{7.070}\\ \textbf{± 0.03}\end{tabular} &
    \begin{tabular}[c]{@{}l@{}}\textbf{-34.84}\\ \textbf{± 0.12}\end{tabular} &
    \begin{tabular}[c]{@{}l@{}}\textbf{-2.796}\\ \textbf{± 0.01}\end{tabular} &
    \begin{tabular}[c]{@{}l@{}}\textbf{0.320}\\ \textbf{± 0.01}\end{tabular} &
    \begin{tabular}[c]{@{}l@{}}\textbf{-2.765}\\ \textbf{± 0.03}\end{tabular} &
    \begin{tabular}[c]{@{}l@{}}\textbf{-1.064}\\ \textbf{± 0.01}\end{tabular} &
    \textbf{1.1}\\
    \bottomrule
    \end{tabular}
    \end{center}
    \normalsize
    \vspace{-4mm}
\end{table}
\begin{table}[t]
    \caption{Cumulative wall-clock time for 15 iterations and averaged over 10 repeated runs (log10 second).}
    \small
    \label{tab:time}
    \centering
    \begin{center}
    \begin{tabular}{p{17mm}p{7mm}p{7mm}p{7mm}p{7mm}p{7mm}p{7mm}p{7mm}p{5mm}cc|c}
    \toprule
    baselines & 
    Ackley &
    \begin{tabular}[c]{@{}l@{}}Rosen-\\ brock\end{tabular} &
    \begin{tabular}[c]{@{}l@{}}Hart-\\ mann\end{tabular} &
    Shekel &
    Pest &
    MaxSat &
    Ising & SVM & 
    \begin{tabular}[c]{@{}l@{}}Mal-\\ aria\end{tabular} &
    \begin{tabular}[c]{@{}l@{}}Sol-\\ vent\end{tabular} &
    \begin{tabular}[c]{@{}l@{}}Mean\\ rank\end{tabular}
    \\
    \midrule
    Random & -1.92	& -1.96 & -1.26 & -1.17 & -1.92 & -1.89 & -1.64 & 0.82 & 1.40 & 1.49 &
    -
    \\
    TS &
    2.71	& 3.10 & 2.79 & 2.86 &	3.00 &	3.70 & 3.22 & 3.36 & 2.71 & 2.85 &
    3.1
    \\
    decoupled TS &
    \textbf{2.20}	& \textbf{2.04} & \textbf{2.01} & \textbf{2.04} & 3.17 & 3.22 & 3.65 & 3.90 & - & - &
    2.6
    \\
    DPP-TS &
    4.85	& 4.56 & 4.35 & 4.62 & 5.67 & 4.49 & 4.73 & 4.73 & - & - &
    7.4
    \\
    TurBO &
    3.42 &	3.06 & 2.12 & 3.07 &	\textbf{2.91} & 2.97 & 3.45 & 3.58 & - & - &
    3.3
    \\
    GIBBON &
    4.92 & 4.18 & 3.71 & 3.52 & 3.72 & 4.71 & 4.25 & 4.41 &  - & - &
    6.8
    \\
    Hallucination &
    4.52 &	4.09 & 4.42 & 3.68 & 4.68 & 4.75 & 4.14 & 5.05 & - & - & 
    7.4
    \\
    LP &
    5.50 &	5.48 & 5.23 & 4.78 & 3.84 & 5.48 & 5.10 & 4.53 & - & - &
    8.5
    \\
    SOBER-TS &
    3.10 & 3.43 & 3.16 & 3.17 & 3.30 & 4.01 & 3.20 & 3.21 & - & - &
    4.1
    \\
    SOBER-LFI &
    2.58 & 2.19 & 2.08 & 2.65 & 2.99 & \textbf{2.96} & \textbf{2.28} & \textbf{2.31} & \textbf{2.43} & \textbf{2.35} & \textbf{1.5} \\
    \bottomrule
    \end{tabular}
    \end{center}
    \normalsize
    \vspace{-4mm}
\end{table}
\begin{wrapfigure}[22]{R}{0.48\textwidth}
  \centering
  \vspace{-4mm}
  \includegraphics[width=0.48\textwidth]{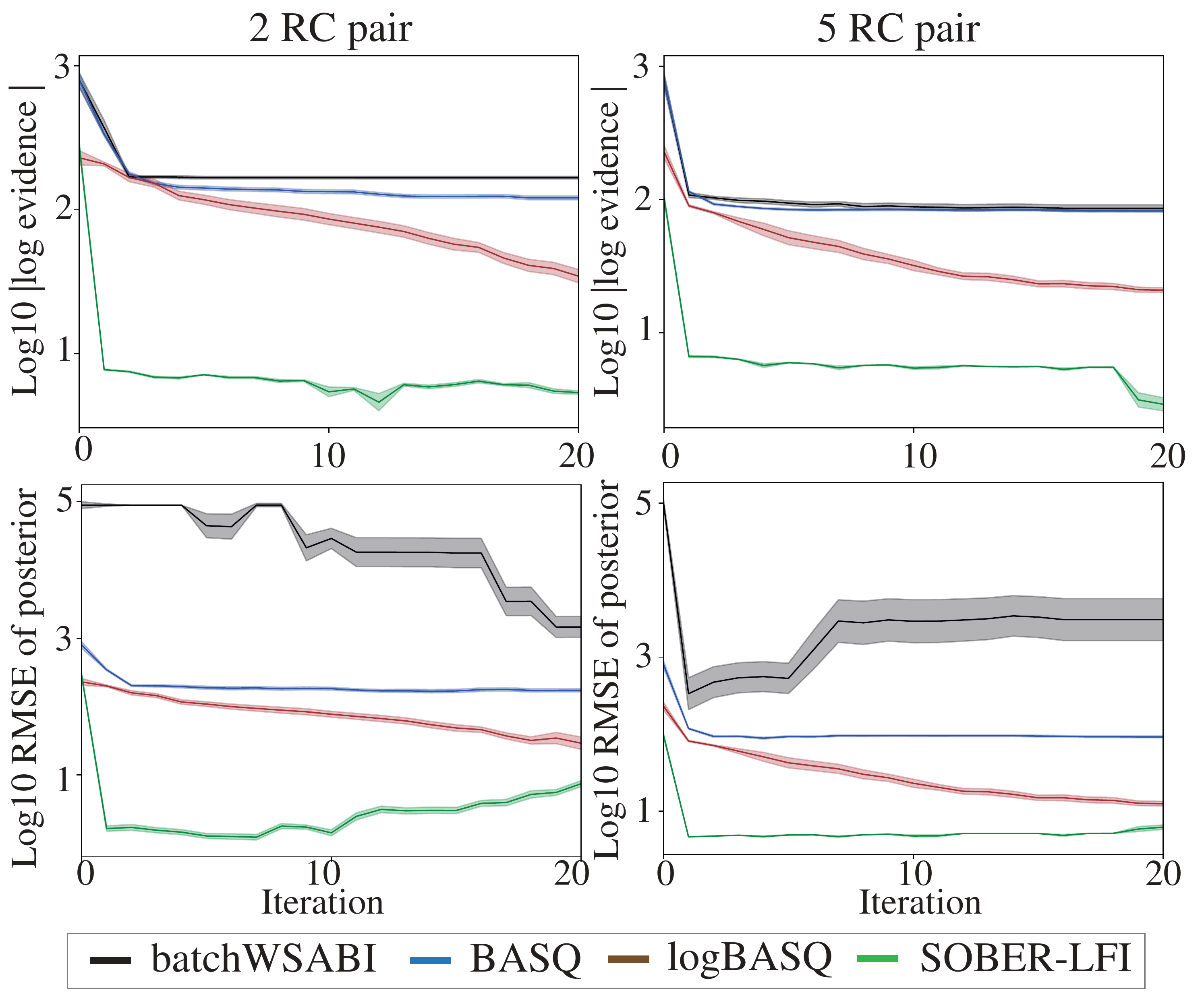}
  \caption{Batch \gls{bq} baseline comparisons across 2 real-world simulation-based inference tasks. \emph{Top}: Log of log evidence. \emph{Bottom}: Log RMSE of posterior as function of iterations. Lines and shaded area denote mean $\pm$ 1 std. error. SOBER-LFI consistently outperforms all three baselines.}
  \label{fig:expbq}
\end{wrapfigure}
Tables \ref{tab:conv} and \ref{tab:time} illustrate the convergence and wall-clock time of sampling overhead at 15th iteration, respectively. As mean rank shows, \gls{sober}-\gls{lfi} outperforms all 8 baselines on 9 out of 10 experiments as well as maintaining the smallest average overhead. Diversified but squeezed batching with \gls{sober}-\gls{lfi} is demonstrated to work well over various levels of multimodal and noisy functions over continuous, discrete, and mixed spaces. 
\gls{sober}-\gls{lfi} was only outperformed on the unimodal noiseless Rosenbrock function, which is in favour of hill-climbing algorithms like \gls{ts} and TurBO. Still, \gls{sober}-\gls{lfi} takes second place, showing that updating $\pi$ can efficiently squeeze the sampling region around the global maximum. In drug discovery tasks, \gls{sober}-\gls{lfi} showed fast convergence in both optimisation and computation, where most algorithms do not apply. Particularly, the solvent dataset clearly exemplifies the stuck behaviours (see Supp. F for learning curves). \gls{ts} converges fast in the early stage, but it can not get out of the local maxima, resulting in a final regret being identical to random search, whereas \gls{sober}-\gls{lfi} does not suffer.

Figure \ref{fig:expbq} illustrates that \gls{sober} also outperformed the \gls{bq} baselines.
While the original \gls{basq} over-explores the prior distribution and shows plateaus, log\gls{basq} alleviates this behaviour via log-warp modelling. Nonetheless, \gls{sober} showed significantly faster convergence than all competitors in both posterior and evidence inference of all tasks by squeezing $\pi$ toward the posterior mode.\looseness=-1

We used type-II \gls{mle} \gls{gp} throughout the experiments (no auxiliary algorithms from Sec.~\ref{sec:aux} except simulation-based inference). Supp. F further illustrates the effect of \gls{af}, batch size $n$, and hyperparameters $(N, M)$. Amongst the \glspl{af}, information-theoretic \glspl{af} (MES and GIBBON) can boost the convergence rate than the \gls{lfi} \gls{af} with a negligible overhead increase. The hyperparameters $N$, $M$ are quite intuitive: the discretisation accuracy of the input $x$ and function $f$ spaces, respectively. Unsurprisingly, the larger these values become, the faster the convergence becomes but the larger the overhead is. Our default values $(N=20000, M=500)$ are competitive throughout the experiments. These values can be adjusted to the cost of queries \cite{adachi2022fast} \footnote{see guidelines in \cite{adachi2022fast}}.
Moreover, the larger the batch size $n$ becomes, the faster the convergence\footnote{We emphasise the convergence acceleration in the large batch is not typically achievable with other baseline methods, such as GIBBON (See Figure 6 in \cite{moss2021gibbon}).}. The ablation study shows that each component (temporary likelihood $\pi$, the iterative $\pi$ update, and the objective-\gls{rchq}) contributes to faster convergence. Additionally, \gls{fbgp} with quadrature distillation can accelerate convergence, especially in noisy functions, while maintaining the overhead is competitive enough to the baselines.

\section{Discussion}
We introduced a hallucination-free approach, SOBER, capable of scalable batching for both \gls{bo} and \gls{bq}. We identified three problems of existing batch \gls{bo} methods: batch diversity, batch size scalability, and combinatorial explosion in innumerable discrete/mixed spaces. The batch \gls{bq} reformulation can make the batch selection more diversified and parallelisable, the objective-\gls{rchq} offers a scalable batch selection solver, and updating $\pi$ avoids the combinatorial explosion via squeezing the feasible region. Updating $\pi$ also solves the over-exploration issue in batch \gls{bq}. This work is extended to the constrained optimization \cite{adachi2023domain}.
Two limitations of \gls{sober} are that it is not suitable for asynchronous settings and that the algorithm cannot be distributed to each node in a computer cluster such as in \cite{hernandez2017parallel}. Applicability to high-dimensional \gls{bo} is also an open problem as efficient sampling from the posterior over the maximiser is difficult. This can be true also for low-dimensional embeddings, such as those produced by the GPLVM model \cite{lawrence2003gaussian}.
However, for some embeddings, such as linear embeddings or VAEs, this sampling can be done efficiently. The reason why the empirical measure with random samples can work surprisingly well has not been fully understood. Study of random convex hulls and related hypercontractivity \cite{hayakawa2023estimating,hayakawa2022hypercontractivity} explains it in a slightly different setting. Future theoretical analysis would give a better understanding and development.\looseness=-1

\section*{Acknowledgments}
We thank Leo Klarner for the insightful discussion of Bayesian optimistaion for drug discovery, Samuel Daulton, Binxin Ru, and Xingchen Wan for the insightful discussion of Bayesian optimisation for graph and mixed space, Ondrej Bajgar for his helpful comments about improving the paper. Masaki Adachi was supported by the Clarendon Fund, the Oxford Kobe Scholarship, the Watanabe Foundation, and Toyota Motor Corporation. Harald Oberhauser was supported by the DataSig Program [EP/S026347/1] and the Hong Kong Innovation and Technology Commission (InnoHK Project CIMDA). Martin Jørgensen was supported by the Carlsberg Foundation.

\bibliographystyle{plainurl}
\sloppy
\bibliography{references}

\appendix
\newpage
\section{Background}
\subsection{Bayesian Optimisation}
\gls{bo} is an algorithm for solving the black-box global optimisation problems defined by:
\begin{equation}
x^*_\text{true} = \arg \max_x f_\text{true}(x), \label{eq:BO}
\end{equation}
where $x^*_\text{true}$ is the true global maximum of the true black-box function $f_\text{true}$. The underlying assumptions here are:
\begin{compactenum}
    \item $f_\text{true}$ is a black-box function, and we do not know the function information except for querying the values at certain locations (oracles).
    \item We can query multiple locations at the same time without additional overhead, and a larger batch size is desirable for faster convergence in wall-clock time.
\end{compactenum}
With the given assumption, the desiderata are:
\begin{compactenum}
    \item The queries are expensive, and we wish to minimise the number of queries for fast convergence.
    \item The total overhead of the batch acquisition algorithm should be smaller for fast computation.
\end{compactenum}
For sample-efficient global optimisation, \gls{bo} typically utilities a surrogate \gls{gp} model $f$ of $f_\text{true}$. We set zero mean \gls{gp} as a prior over function space, and set Gaussian likelihood. As both are Gaussian and conjugate, the predictive posterior with noisy observations has the closed-form as Gaussian, denoted by:
\begin{subequations}
\begin{align}
    &P(f \mid \textbf{D}_\text{obs}, x) = \mathcal{GP}\Big(f; m(x), C(x,x^\prime) \Big),\label{eq:noiseless}\\
    &m(x) = K(x, \textbf{X}_\text{obs}) \Big[ K(\textbf{X}_\text{obs}, \textbf{X}_\text{obs}) + \sigma^2 \textbf{I} \Big]^{-1} \textbf{y}_\text{obs}, \label{eq:pred_mean}\\
    &C(x,x^\prime) = K(x, x^\prime) - K(x, \textbf{X}_\text{obs}) \Big[ K(\textbf{X}_\text{obs}, \textbf{X}_\text{obs}) + \sigma^2 \textbf{I} \Big]^{-1} K(\textbf{X}_\text{obs}, x^\prime), \label{eq:pred_cov}
\end{align}
\end{subequations}
where $\textbf{D}_\text{obs} \!=\! (\textbf{X}_\text{obs}, \textbf{y}_\text{obs})$ is the observed dataset, $\textbf{y}_\text{obs} \!:=\! f_\text{true}(\textbf{X}_\text{obs})$ are the oracles queried in parallel, $m(\cdot)$ and $C(\cdot, \cdot)$ are the mean and covariance of the predictive posterior, $K(\cdot, \cdot\mid\theta)$ is the kernel, $\theta$ is the kernel hyperparameters, $\textbf{I}$ is the identity matrix, $\sigma^2_n$ is the variance of Gaussian noise, and $x$ and $x^\prime$ are the locations where we wish to predict, $x, x^\prime \notin \textbf{X}_\text{obs}$.

$m$ and $C$ guide our beliefs toward the region where the true global maximum $x^*_\text{true}$ possibly locates.
Such a guiding mechanism is obtained through maximising an \gls{af}, which selects the next query via its maximisation.
There are several types of \glspl{af}, such as \gls{ei} \cite{jones1998efficient}, \gls{ucb} \cite{srinivas2009gaussian}, information-theoretic \glspl{af} \cite{villemonteix2009informational, hennig2012entropy, wang2017max, hvarfner2022joint}, \gls{fbgp}-based \glspl{af} \cite{houlsby2011bayesian, seung1992query, riis2022bayesian}. More sample-efficient \gls{af} tends to be more computationally expensive.

Queried observations $\textbf{D}_\text{obs}$ serially update the \gls{gp} surrogate model $f$ so it can predict the output of $f_\text{true}$ more accurately. When updating \gls{gp} with given $\textbf{D}_\text{obs}$, \gls{gp} hyperparameters $\theta$ are also updated. There are two ways of updating hyperparameters; type-II \gls{mle} and \gls{fbgp}. While type-II \gls{mle} is the point estimation of optimal hyperparamter in terms of the marginal likelihood of the \gls{gp}, \gls{fbgp} estimates the hyperposterior, typically performed by \gls{mcmc}, then represent the predictive posterior as the ensemble of \gls{gp} with the hyperparameters randomly sampled from hyperposterior.

\subsection{Bayesian Quadrature}
\gls{bq} is an algorithm for evaluating integrals given by:
\begin{equation}
Z_\text{true} = \int f_\text{true}(x) \text{d}\pi(x), \label{eq:bq}
\end{equation}
where $f_\text{true}$ is the black-box function we wish to integrate against a known probability measure $\pi$.
The difference from \gls{bo} is the objective being integration, not global optimisation. The integration problem is widely recognised in statistical learning: expectations, variances, marginalisation, ensembles, Bayesian model selection, and Bayesian model averaging. \gls{bq} is, like \gls{bo}, solved by \gls{gp}-surrogate-model-based active learning. The batch acquisition methods are also shared with batch \gls{bo}. The methodological differences are:
\begin{compactenum}
    \item \gls{bq} typically assumes a specific kernel to make the integration analytical (e.g.~RBF kernel).
    \item While \gls{bo} requires to approximate the black-box function only in the vicinity of the global optimum, \gls{bq} needs to approximate the whole region of interest defined by the probability measure $\pi$.
\end{compactenum}
Thus, \gls{bq} is a purely explorative algorithm, and the uncertainty sampling \gls{af} is often applied.

The classic method to estimate the integral exploits Gaussianity. Let $\pi$ be multivariate normal distribution $\pi(x) = \mathcal{N}(x; \mu_\pi, \boldsymbol\Sigma_\pi)$, and the kernel $K$ be RBF kernel, which can be represented as Gaussian $K(\textbf{X}_\text{ob}, x) = v \sqrt{|2\pi \textbf{W}|} \mathcal{N}(\textbf{X}_\text{ob}; x, \textbf{W})$, where $v$ is kernel variance and $\textbf{W}$ is the diagonal covariance matrix whose diagonal elements are the lengthscales of each dimension. As the product of two Gaussians is a Gaussian, the integrand becomes a Gaussian and its integral has the closed form, as such:
\begin{align}
\int m(x)\pi(x)dx 
&= v \left[ \int \mathcal{N}(x; \textbf{X}_\text{obs}, \textbf{W}) \mathcal{N}(x; \mu_\pi, \boldsymbol\Sigma_\pi) dx \right]^\top
K(\textbf{X}_\text{obs}, \textbf{X}_\text{obs})^{-1} \textbf{y}_\text{obs},\\ 
&= v \left[ \int \mathcal{N}(x; \textbf{X}_\text{obs}, \textbf{W}) \mathcal{N}(x; \mu_\pi, \boldsymbol\Sigma_\pi) dx \right]^\top \boldsymbol\omega,\\ 
&= v \mathcal{N}(\textbf{X}_\text{obs}; \mu_\pi, \textbf{W}+\boldsymbol\Sigma_\pi)^\top \boldsymbol\omega,
\end{align}
where $\boldsymbol\omega := K(\textbf{X}_\text{obs}, \textbf{X}_\text{obs})^{-1} \textbf{y}_\text{obs}$. As such, the integration of \gls{gp} over the measure $\pi$ is analytical. The more accurately the \gls{gp} can approximate the true function $f_\text{true}$, the more accurately the above integration estimation approximates.

As such, classical \glspl{bq} have additional limitations on prior and kernel selections. To make the integration closed-form, the prior needs to be uniform or Gaussian, and the kernel also needs to be limited selection (e.g. RBF kernel, see Table 1 in \cite{briol2019probabilistic}).

\section{Details on Related Works}
\subsection{Compatibility}
The compatibility of each batch \gls{bo} method is summarised in Table \ref{tab:comparison}.

\paragraph{Hallucination}
Hallucination \cite{azimi2010batch} tackled batch \gls{bo} by simulating sequential process by putting “fantasy” oracles estimated by \gls{gp}, translating batch selection into sequential problem.  Hallucination is successful in low batch size $n$, but not scalable. Even a single iteration of \gls{af} maximisation is not trivial due to non-convexity, but they repeat this over $n$ times and produce prohibitive overhead. For discrete and mixed space, maximizing \gls{af} requires enumerating all possible candidates. However, the higher the dimension and larger the number of categorical classes, the more infeasibly large the combination becomes  (combinatorial explosion).

\paragraph{Local Penalisation (LP)}
\gls{lp} \cite{gonzalez2016batch}, simulates only \gls{af} shape change, without fantasy oracles, by penalising \gls{af} assuming Lipschitz continuity. This succeeds in speeding up the hallucination algorithm. However, the principled limitations are inherited (combinatorial explosion). Large batch sizes are also not applicable because maximising \gls{af} still produces large overhead. This is because maximising \gls{af} is typically computed by multi-start optimiser, but the number of random seeds need to increase dependent on the number of dimensions and multimodality of true function. This optimiser also does not guarantee to be globally maximised, which contradicts the assumption of \gls{af} (only optimal if it is globally maximised.). Furthermore, Lipschitz continuity assumption limits its applicable range to be only for continuous space. 

\paragraph{Thompson Sampling (TS)}
\gls{ts} \cite{hernandez2017parallel} is a random sampling method of $P(x^* \mid \textbf{D}_\text{obs})$ by maximising the function samples drawing from predictive posterior in Eq. \eqref{eq:noiseless}. Due to its random sampling nature, exactly maximising the function samples are not strict relative to hallucination or \gls{lp}. Thus, in practice, \gls{ts} is typically done with taking argmax of function samples amongst the candidates of random samples over input space. This two-step sampling nature (random samples over input space $\rightarrow$ subsamples with argmax of random function samples) allows us for domain-agnostic \gls{bo}. However, this scheme itself is a type of \gls{af}, so other \gls{af} is not naïvely supported. Moreover, due to the random sampling nature, the selected batch samples are not sparsified to efficiently explore uncertain regions.

\paragraph{Decoupled TS}
Decoupled \gls{ts} \cite{wilson2020efficiently} achieved faster but more diversified sampling than \gls{ts} by separating out the prior from the data. Only this method and \gls{sober} sparsifies the samples via $f$-space (function space), which allows for general-purpose sparsifications. However, this does not guarantee to be $f$-space sparsification leads to better sparcification in $x$-space. Figure \ref{fig:demo} visualises that decoupled \gls{ts} can provide more diversified samples than \gls{ts}, however, it still suffered from being stuck in local minima. Also, \glspl{af} are not custamisable.

\paragraph{Determinantal Point Process (DPP)}
\gls{dpp}-\gls{bo} \cite{kathuria2016batched} proposed diversified sampling using \gls{dpp} via maximising the determinant of Gram matrix of the selected batch samples. However, this maximisation problem in general is NP-hard and the best known sampling method is a rejection sampling, which produces significant overhead. Thus, large batch sizes are prohibitive. This method also cannot support discrete and mixed spaces due to combinatorial explosion.

\paragraph{DPP-TS}
\gls{dpp}-\gls{ts} tried to take the best of both worlds of \gls{dpp} and \gls{ts}. Thanks to the randomised sampling of \gls{ts}, this method can apply to discrete and mixed spaces, avoid combinatorial explosion, and is faster than original \gls{dpp}. Still, the computation is much slower than non-DPP-based alternatives.

\paragraph{MC-SAA}
MC-SAA \cite{balandat2020botorch} adopts \gls{qmc}-based \gls{af} approximation, which achieves faster computations with a deterministic function sampler. Still, underlying strategy of \gls{af} maximisation is shared with classic methods (hallucination, \gls{lp}), hence the limitations are also shared. Thus, \gls{af} maximisation requires enumerating all combinations of possible candidates, easily becomes combinatorial explosion. Moreover, this method is only applicable to those \gls{mc} approximation can be applied, so information-theoretic \gls{af} cannot be applied.

\paragraph{GIBBON}
GIBBON \cite{moss2021gibbon} proposed diversified sampling methods using a specific \gls{af} (lower bound approximation of max-value entropy search \cite{wang2017max}). Querying large batch sizes requires enumerating all possible permutations of \textit{both batch samples and discrete variables}. This requirement introduces a combinatorial explosion. Moreover, even in continuous variables, GIBBON becomes less diversified for large batch size (See Figure 6 in \cite{moss2021gibbon}).

\paragraph{TurBO}
TurBO \cite{eriksson2019scalable} introduced multiple local \glspl{bo} bounded with trust regions, and allocates batching budgets based on \gls{ts}. This succeeded in scalable batching via maintaining local \glspl{bo} that are compact, via shrinking trust regions, based on heuristics with many hyperparameters. Selecting hyperpameters is non-trivial and TurBO cannot apply to discrete and non-Euclidean space, for which kernels do not have lengthscale hyperparameters for the trust region update heuristic (e.g. Tanimoto kernel for drug discovery \cite{ralaivola2005graph}).

\subsection{Sampling Complexity}
\begin{table}
  \caption{Comparion with batch BO algorithms.
  The scalability to large batch size is defined by the computational complexity that is smaller or equivalent complexity than \gls{ts}.}
  \small
  \label{tab:complexity}
  \centering
  \begin{center}
  \begin{tabular}{lllll}
    \toprule
    Batch BOs &
    \begin{tabular}{@{}l@{}}large \\ batch\end{tabular} &
    \begin{tabular}{@{}l@{}}mixed \\ space\end{tabular} &
    \begin{tabular}{@{}l@{}}No combinatorial \\ explosion \end{tabular} &
    sampling complexity
    \\
    \midrule
    Hallucination \cite{azimi2010batch} & \ding{53} & \ding{51} & \ding{53} &  $n (N C_\text{acq} + C_\text{update} + N \log N)$ \\
    Local penalisation \cite{gonzalez2016batch} & \ding{51} & \ding{53} & \ding{53} & $n (N C_\text{acq} + N \log N)$ \\
    TS \cite{hernandez2017parallel} & \ding{51} & \ding{51} &  \ding{51} & $n (N C_\text{func} + N \log N)$ \\
    Decoupled TS \cite{wilson2020efficiently} & \ding{51} & \ding{51} &  \ding{51} & $n (N M C_\varphi + N \log N)$ \\
    DPP 
    \cite{kathuria2016batched} & \ding{53} & \ding{51} & \ding{53} & $n^3 N C_\text{reject}$\\
    DPP-TS \cite{nava2022diversified} & \ding{51} & \ding{51} & \ding{51} & $ I_\text{MCMC} (N C_\text{func} + n^3)$ \\
    MC-SAA \cite{balandat2020botorch} & \ding{53} & \ding{51} & \ding{53} & $P^N_n C_\text{acq} + P^N_n \log P^N_n$ \\
    GIBBON \cite{moss2021gibbon} & \ding{53} & \ding{51} & \ding{53} & $ P^N_n C_\text{acq} + P^N_n \log P^N_n$ \\
    SOBER (Ours) & \ding{51} & \ding{51} & \ding{51} & $ N M C_\varphi + N C_\text{acq} + n^3 \log (N/n)$ \\
    \bottomrule
  \end{tabular}
  \end{center}
  \normalsize
\end{table}

Table \ref{tab:complexity} summarises the computational complexity of each baseline method for sampling. The sampling complexity shows the upper bounds of $N$ discrete candidates. Notation remarks: $C_\text{reject}$ is the cost of rejection sampling from \gls{dpp}, $C_\text{acq}$ is the cost of evaluating the \gls{af} at a point, $C_\text{update}$ is the cost of updating the \gls{gp}, including the Gram matrix inversion and hyperparameter optimisation, $C_\text{func}$ is the cost of evaluating a sampled function at a point, $I_\text{MCMC}$ is the number of \gls{mcmc} iterations until convergence, $M C_\varphi$ is the cost of evaluating $M$ components of the kernels approximated by Nystr{\"o}m method or random Fourier feature, $P^N_n := n! / (N - n)! \gg nN$ is the total number of all permutations, $N$ is the number of candidate points, $n$ is the batch size. The computational costs are in order of $C_\text{reject} \gg C_\text{update} \gg C_\text{func} \geq C_\text{acq} > M C_\varphi$, and the number of samples are in order of $N > M > n$, $I_\text{MCMC} \gg n$. The empirical comparison can be seen in Table \ref{tab:time}. Our method was the fastest.

\section{Quadrature Distillation} \label{sec:qd}
\subsection{Fast Fully Bayesian Gaussian Process}
\subsubsection{Existing Method: MCMC}
While \gls{bo} works well with a single set of optimised hyperparameters (type-II \gls{mle}) on most functions, some noisy cases such as drug discovery shown in Figure \ref{fig:qm9} requires \gls{fbgp} modelling. \gls{gp} marginalisation is typically done with \gls{mc} integration via sampling from hyperposterior $\Pi$. The hyperposterior $\Pi$ is given by:
\begin{align}
    \Pi(\theta) &:= p(\theta \mid \textbf{D}_\text{obs}),\\
    &= \frac{\int p(\textbf{y}_\text{obs}\mid f) p(f\mid\theta, \textbf{X}_\text{obs}) p(\theta) \text{d} f}{\iint p(\textbf{y}_\text{obs}\mid f) p(f\mid\theta, \textbf{X}_\text{obs}) \text{d} f \text{d} p(\theta)} ,\\
    &= \frac{L(\theta) \Pi^\prime(\theta)}{\int L(\theta) \text{d} \Pi^\prime(\theta)},\\
    &= \frac{L(\theta) \Pi^\prime(\theta)}{\mathcal{L}_\text{hyper}},
\end{align}
where
\begin{align}
    \Pi^\prime(\theta) &:= p(\theta),\\
    L(\theta) &:= \int p(\textbf{y}_\text{obs}\mid f) p(f\mid\theta, \textbf{X}_\text{obs}) \text{d} f,\\
    &= \mathcal{N}\Big( \textbf{y}_\text{obs}; m(\textbf{X}_\text{obs}\mid\theta), C(\textbf{X}_\text{obs}, \textbf{X}_\text{obs}\mid\theta) \Big) ,\label{eq:ml} \\ 
    \mathcal{L}_\text{hyper} &= \int L(\theta) \text{d} \Pi^\prime(\theta),
\end{align}
$\Pi^\prime(\theta)$ is the hyperprior, $L$ is the marginal likelihood, $\mathcal{L}_\text{hyper}$ is the marginal hyperlikelihood.
And the predictive posterior for the test inputs $x$ can be approximated with \gls{mcmc} with $M$ samples as
\begin{align}
    p(y \mid \textbf{D}_\text{obs}, x, x^\prime) 
    &:= \iint p(y \mid f, x, x^\prime)p(f \mid \theta, \textbf{D}_\text{obs}) \text{d} f \text{d} p(\theta \mid \textbf{D}_\text{obs}),\\
    &= \int p(y \mid \theta, \textbf{D}_\text{obs}, x, x^\prime) \text{d} p(\theta \mid \textbf{D}_\text{obs}),\\
    &= \int L(x \mid \theta) \text{d} \Pi(\theta),\\
    &\approx \frac{1}{M} \sum_{m=1}^M L(x \mid \theta_m)
\end{align}
where
\begin{align}
    L(x \mid \theta) &:= \mathcal{N}\Big(y; m(x \mid \theta), C(x, x^\prime \mid \theta) \Big)\\
   \theta_m &\sim \Pi
\end{align}
However, its convergence rate $\mathcal{O}(1/\sqrt{n})$ is poor, requiring thousands of ensemble GP models for marginalisation. This significantly slows down batch \gls{bo} computation as BASQ requires such an ensemble kernel. Hence, we introduce \gls{qd} trick for faster marginalisation by using the smaller set of weighted hypersamples selected by the quadrature, distilling random hypersamples from hyperprior. 

\subsubsection{MCMC-based Quadrature Distillation}
The easiest to implement is to apply existing \gls{mcmc} scheme to sample from hyperposterior $\Pi$. Let $\boldsymbol\theta_\text{cand}$ be the candidates sampled from $\Pi$ via \gls{mcmc}, $\boldsymbol\theta_\text{cand} \sim \Pi$. Then, simply applying \gls{rchq} with an arbitrary kernel yields the small subset of $(\textbf{w}_\text{QD}, \boldsymbol\theta_\text{QD})$. We refer this weighted subset of hypersamples to \gls{qd} samples, which are the approximation of $\Pi$. Intuitively, this can be understood as the weighted samples can more sample-efficiently approximate the distribution than random samples that approximate the distributions with frequencies. With the \gls{qd} samples, we can approximate the marginal predictive posterior of \gls{fbgp}, given by:
\begin{align}
    p(y \mid \textbf{D}_\text{obs}, x, x^\prime) &\approx \textbf{w}_\text{QD}^\top L(x \mid \boldsymbol\theta_\text{QD}), \label{eq:mcmcqd}\\
    \begin{split}
    m_\text{QD}(x) &= \int m(x\mid\Theta) \text{d} \Pi(\theta),\\
    &\approx \textbf{w}_\text{QD}^\top  m(x\mid\boldsymbol\theta_\text{QD}),
    \end{split}\\
    \begin{split}
    V_\text{QD}(x) &= \int C(x, x\mid\theta) \text{d} \Pi(\theta),\\
    &\approx \textbf{w}_\text{QD}^\top \left[ C(x, x\mid\boldsymbol\theta_\text{QD}) + m^2(x\mid\boldsymbol\theta_\text{QD}) \right] - m_\text{QD}^2(x).
    \end{split}\\
    \begin{split}
    C_\text{QD}(x, x^\prime) &= \int C(x, x^\prime\mid\theta) \text{d} \Pi(\theta),\\
    &\approx \sum_i^H w_\text{i, QD} \left[ \left(m(x\mid\theta_\text{i, QD}) - m_\text{QD}(x) \right)^T  \left( m(x^\prime\mid\theta_\text{i, QD}) - m_\text{QD}(x^\prime) \right) \right],
    \end{split}
\end{align}
where $\textbf{w}_\text{QD} \in \mathbb{R}^H$ is the \gls{qd} weights, $\boldsymbol{\theta}_\text{QD} \in \mathbb{R}^{D \times H}$ is the \gls{qd} samples, $D$ is the number of types of hyperparameters (e.g. lengthscale, variance), and $H$ is the number of \gls{qd} samples, which is much smaller than the number of \gls{mcmc} samples $M$, $H \ll M$. Thus, we can estimate the marginal predictive posterior of \gls{fbgp} as the small set of weighted hypersamples via \gls{qd}. As all the above marginalisation shares the same signed measure $\Pi$, so the quadrature with \gls{qd} is good approximation.

\subsubsection{BQ-based Quadrature Distillation}
\gls{mcmc}-based \gls{qd} suffices for most cases, however, \gls{mcmc} produces significant overhead, so we wish to avoid this sampling scheme for quick estimation. Thus we adopt \gls{bq}-based quadrature distrillation. 
First, we apply \gls{bq} by placing \gls{gp} on hyperlikelihood $L(\theta)$,
\small
\begin{align}
\begin{split}
    \Pi(\theta) &= \frac{L(\theta) \Pi^\prime(\theta)}{\int L(\theta) \text{d} \Pi^\prime(\theta)},\\
    &\approx 
    \frac{ \mathbb{E}\Big[ p(\ell \mid \theta, \textbf{L}_\text{obs}, \boldsymbol\theta_\text{obs}) \Big] \Pi^\prime(\theta)}{ \int \mathbb{E}\Big[ p(\ell \mid \theta, \textbf{L}_\text{obs}, \boldsymbol\theta_\text{obs}) \Big] \text{d} \Pi^\prime(\theta)},\\
    &\approx 
    \frac{m_\text{hyper}(\theta)\Pi^\prime(\theta)}{ \int m_\text{hyper}(\theta) \text{d} \Pi^\prime(\theta)},\\
    &=
    \frac{m_\text{hyper}(\theta)\Pi^\prime(\theta)}{\left[
    \int K_\text{hyper}(\theta, \boldsymbol\theta_\text{obs}) K_\text{hyper}(\boldsymbol\theta_\text{obs}, \boldsymbol\theta_\text{obs})^{-1} \text{d}
    \Pi^\prime(\theta) \right]
    \textbf{L}_\text{obs}},\\
    &= \frac{m_\text{hyper}(\theta)\Pi^\prime(\theta)}{\textbf{w}_\text{BQ}^{\prime \top} \textbf{L}_\text{obs}},
\end{split} \label{eq:hyperposterior}
\end{align}
\normalsize
where
\small
\begin{subequations}
\begin{align}
    P(\ell \mid \theta, \textbf{L}_\text{obs}, \boldsymbol\theta_\text{obs}) &\sim \mathcal{GP}(\ell; m_\text{hyper}(\theta), C_\text{hyper}(\theta,\theta^\prime)),\\
    \textbf{w}_\text{BQ}^\prime &:=
    \int K_\text{hyper}(\theta, \boldsymbol\theta_\text{obs}) K_\text{hyper}(\boldsymbol\theta_\text{obs}, \boldsymbol\theta_\text{obs})^{-1} \text{d}
    \Pi^\prime(\theta), \label{eq:BQweights}\\
    \textbf{L}_\text{obs} &:= L(\boldsymbol\theta_\text{obs}).
\end{align}
\end{subequations}
\normalsize
Here, we place hyper-\gls{gp} on the marginal likelihood $L$ defined by Eq.\ (\ref{eq:ml}). Importantly, this measure is the hyperprior $\Pi^\prime$, not the hyperposterior $\Pi$ like the \gls{mcmc}-based cases. We draw hypersamples $\boldsymbol\theta_\text{obs} \in \mathbb{R}^{D \times M}$, from the hyperprior $\Pi^\prime(\cdot) := P(\theta)$. Then, we evaluate the marginal likelihood $\textbf{L}_\text{obs} = \mathcal{L}(\boldsymbol\theta_\text{obs})$ in parallel. We select multivariate normal distribution for hyperprior $\Pi^\prime(\theta) := \mathcal{N}(\theta; \mu_\text{hyper}, \boldsymbol\Sigma_\text{hyper})$ based on \cite{malkomes2016bayesian}, and Gaussian kernel for hyper-\gls{gp}. Then, weights $\textbf{w}_\text{BQ}^\prime$ in Eq.\ (\ref{eq:BQweights}) become analytical:
\begin{align}
    \begin{split}
    \textbf{w}_\text{BQ}^\prime &:= \int K_\text{hyper}(\theta, \boldsymbol\theta_\text{obs}) K_\text{hyper}(\boldsymbol\theta_\text{obs}, \boldsymbol\theta_\text{obs})^{-1} \text{d}
    \Pi^\prime(\theta),\\
    &= v\sqrt{|2\pi\textbf{W}|}
    \left[\int
    \mathcal{N}(\theta; \boldsymbol\theta_\text{obs}, \textbf{W})\mathcal{N}(\theta; \mu_\text{hyper}, \boldsymbol\Sigma_\text{hyper}) \text{d}\theta
    \right]
    K_\text{hyper}(\boldsymbol\theta_\text{obs}, \boldsymbol\theta_\text{obs})^{-1},\\
    &= v\sqrt{|2\pi\textbf{W}|}
    \mathcal{N}(\boldsymbol\theta_\text{obs}; \mu_\text{hyper},  \textbf{W}+\boldsymbol\Sigma_\text{obs})K_\text{hyper}(\boldsymbol\theta_\text{obs}, \boldsymbol\theta_\text{obs})^{-1},
    \end{split}\\
\end{align}

With the given weighted samples $(\textbf{w}_\text{BQ}^\prime, \boldsymbol\theta_\text{obs})$, we can approximate the marginalisation. For instance, the marginal predictive posterior can be approximated as follows:
\small
\begin{align}
\begin{split}
    p(y \mid \textbf{D}_\text{obs}, x, x^\prime) &\approx \int L(x\mid\theta)\frac{m_\text{hyper}(\theta)}{\textbf{w}_\text{BQ}^{\prime \top} \textbf{L}_\text{obs}} \text{d}\Pi^\prime(\theta),\\
    &\approx \textbf{w}_\text{BQ}^\top  L(x\mid\boldsymbol\theta_\text{obs}), \label{eq:margW}
\end{split}
\end{align}
\normalsize
where $\textbf{w}_\text{BQ} := \textbf{w}_\text{BQ}^\prime \odot \textbf{L}_\text{obs} / (\textbf{w}_\text{BQ}^{\prime \top} \textbf{L}_\text{obs})$.
Obviously, Eq. \eqref{eq:margW} is the same approximation with the \gls{mcmc}-based \gls{qd} approximation in Eq. \eqref{eq:mcmcqd}. The difference is the number of hypersamples; the number of hyperprior samples $M$ is larger than that of quadrature distillation.
Therefore, we can further distill the \gls{bq} samples, $ (\textbf{w}_\text{QD}, \boldsymbol{\theta}_\text{QD}) \subset (\textbf{w}_\text{BQ}, \boldsymbol{\theta}_\text{obs})$ via \gls{rchq}. The kernel for \gls{rchq} is the kernel of hyper-\gls{gp}, $K_\text{hyper}$. The kernel hyperparameters are optimised by type-II \gls{mle}. These procedure does not require \gls{mcmc}. Thus, this can offer faster computation.

\subsection{Fast Fully Bayesian Acquisition Functions}
Many \glspl{af}, including the \gls{lfi}, are dependent on not only kernel hyperparameters $\theta$, but also the current maximum $\eta$. To achieve fast marginalisation of \glspl{af}, we need to incorporate $\eta$ as one of arguments in estimation.

\subsubsection{Parabolic Transform for Max-value Estimation}
Estimating the current maximum location conditioned on $\theta$ is computationally challenging. We inherit the parabolic transform of \gls{gp} surrogate model from \cite{ru2018fast}:
\begin{subequations}
\begin{align}
    &f(x\mid\theta) = \eta - \frac{1}{2}g(x)^2,\\
    &\quad\quad\quad := \mathcal{GP}\big( f(x); m(x\mid\theta, \eta), C(x,x\mid\theta) \big), \\
    &g(x\mid\theta) := \mathcal{GP}\big( g(\cdot); m_g(\cdot\mid\theta), C_g(\cdot,\cdot\mid\theta) \big),\\
    &m(x\mid\theta, \eta) := \eta - \frac{1}{2} \left[m_g(x)^2 + C_g(x,x)\right],\\
    &C(x,x^\prime\mid\theta) := \frac{1}{2}C_g(x,x^\prime)^2 + m_g(x)^\top C_g(x,x^\prime) m_g(x^\prime),
\end{align}
\end{subequations}
where $f(\cdot)$ is the surrogate model that approximates $f_\text{true}(\cdot)$, $g(\cdot)$ is the square-root warped \gls{gp} \cite{snelson2003warped} of $f(\cdot)$. The predictive mean $m_g(\cdot)$ and covariance $C_g(\cdot,\cdot)$ of the warped \gls{gp} $g(\cdot)$ are expressed with normal \glspl{gp} in Eqs.\ (\ref{eq:pred_mean}) - (\ref{eq:pred_cov}). The predictive mean $m(\cdot)$ and covariance $C(\cdot,\cdot)$ are approximated via moment-matching \cite{gunter2014sampling}. $\textbf{D}_g = (\textbf{X}_\text{obs}, \textbf{y}_{g, \text{obs}})$ is the observed data for the warped \gls{gp}, and $ \textbf{y}_{g, \text{obs}} := \sqrt{2(\eta - \textbf{y}_\text{obs})}$. Now, $\eta$ becomes a \gls{gp} hyperparameter via $\textbf{y}_{g, \text{obs}}$. All we have to do for \gls{qd} is just replacing the definition of \gls{gp} to this warped \gls{gp}.

\subsubsection{Marginal Expected Improvement Acquisition Function}
The marginal \gls{ei} \gls{af} \cite{jones1998efficient} can be calculated with parabolic-transformed \gls{fbgp} formulation:
\begin{align}
    \alpha_\text{EI}(x) &:= \textbf{w}_\text{QD}^\top \left[(m(x\mid\boldsymbol\theta_\text{QD}) - \boldsymbol\eta) \odot \Phi(Z\mid\boldsymbol\theta_\text{QD})\right] + \textbf{w}_\text{QD}^\top \left[ \sqrt{C(x,x\mid\boldsymbol\theta_\text{QD})} \odot \phi(Z\mid\boldsymbol\theta_\text{QD}) \right]\\
    Z &:= \frac{m(x\mid\boldsymbol\theta_\text{QD}) - \boldsymbol\eta}{\sqrt{C(x,x\mid\boldsymbol\theta_\text{QD})}}
\end{align}
where $\Phi(x), \phi(x)$ are \gls{cdf} and \gls{pdf} of the normal distribution, $\boldsymbol\eta \in \boldsymbol\theta_\text{QD}$ is the distilled max value $\eta$.

\subsubsection{Marginal Upper Confidence Bound Acquisition Function}
The marginal \gls{ucb} \gls{af} \cite{srinivas2009gaussian} can be calculated using \gls{fbgp} formulation without parabolic-transformation:
\begin{align}
    \alpha_\text{UCB}(x) &:= \textbf{w}_\text{QD}^\top m(x\mid\boldsymbol\theta_\text{QD}) + \sqrt{\beta} \textbf{w}_\text{QD}^\top \sqrt{C(x,x\mid\boldsymbol\theta_\text{QD})}
\end{align}
where $\beta$ is the \gls{bo} hyperparameter, usually 0.2 is selected.

\subsubsection{Max-value Entropy Search Acquisition Function}
The \gls{mes} \gls{af} \cite{wang2017max} can be calculated via parabolic-transformed \gls{fbgp} formulation via \texttt{FITBO} formulation \cite{ru2018fast}:
\begin{align}
    \alpha_\text{FITBO}(x) &:= H[p(y\mid\textbf{D}_\text{obs}, x)] - \mathbb{E}_{p(\eta\mid\textbf{D}_\text{obs})}\bigg[
    H[p(y\mid\textbf{D}_\text{obs}, x, \eta)]\bigg],\\
    p(y\mid\textbf{D}_\text{obs}, x) &= \int p(y\mid\textbf{D}_\text{obs}, x, \eta) \text{d} p(\eta\mid\textbf{D}_\text{obs}),\\
    H[p(y\mid\textbf{D}_\text{obs}, x)] &= \int \ln p(y\mid\textbf{D}_\text{obs}, x) \text{d} p(y\mid\textbf{D}_\text{obs}, x),\\
    \mathbb{E}_{p(\eta\mid\textbf{D}_\text{obs})}\bigg[H[p(y\mid\textbf{D}_\text{obs}, x, \eta)]\bigg] &= \int H[p(y\mid\textbf{D}_\text{obs}, x, \eta)] \text{d} p(\eta\mid\textbf{D}_\text{obs}).
\end{align}
FITBO \gls{af} can be discretised via \gls{mc} integration:
\begin{align}
    \alpha_\text{FITBO}(x\mid\textbf{D}_\text{obs}) &:= H\bigg[\frac{1}{M}\sum_i^M p(y\mid\textbf{D}_\text{obs}, x, \theta_i, \eta_i)\bigg] - \frac{1}{2M}\sum_i^M \log [2\pi e(C(x, x\mid\textbf{D}_\text{obs}, \theta_i, \eta_i) + \sigma_{n,i})].
\end{align}
Quadrature distillation can approximate the above \gls{af} as:
\begin{align}
    \alpha_\text{FITBO}(x\mid\textbf{D}_\text{obs}) &\approx H\bigg[\textbf{w}_\text{QD}^\top m(x\mid\boldsymbol\theta_\text{QD})\bigg] - \frac{1}{2} \textbf{w}_\text{QD}^\top \log [2\pi e(C(x, x\mid\textbf{D}_\text{obs}, \boldsymbol\theta_\text{QD}) + \boldsymbol\sigma^2_{n, \text{QD}})].
\end{align}
For faster computation, moment-matching approximation yields the first term as:
\begin{align}
    H\bigg[\frac{1}{M}\sum_i^M p(y\mid\textbf{D}_\text{obs}, x, \theta_i, \eta_i)\bigg]&\approx \frac{1}{2} \log[2\pi e (\mathbb{V}\text{ar}[y]+\sigma^2_{n,i})],\\
    \mathbb{V}\text{ar}[y] &= \frac{1}{M} \sum_i^M \bigg(C(x,x\mid\theta_i) + m^2(x\mid\theta_i)\bigg) - \mathbb{E}[y\theta_i]^2,\\
    \mathbb{E}[y] &= \frac{1}{M} \sum_i^M m(x\mid\theta_i).
\end{align}
Hence,
\begin{align}
    \begin{split}
    \alpha_\text{FITBO}(x\mid\textbf{D}_\text{obs})
    &\approx \frac{1}{2} \log[2\pi e (\mathbb{V}\text{ar}[y]+\textbf{w}_\text{QD}^\top \boldsymbol\sigma^2_{n, \text{QD}})]\\
    &- \frac{1}{2} \textbf{w}_\text{QD}^\top \log [2\pi e(C(x, x\mid\textbf{D}_\text{obs}, \boldsymbol\theta_\text{QD}) + \boldsymbol\sigma^2_{n, \text{QD}})],
    \end{split}\\
    \mathbb{V}\text{ar}[y] &= \textbf{w}_\text{QD}^\top \left( C(x,x\mid\boldsymbol\theta_\text{QD}) + m^2(x\mid\boldsymbol\theta_\text{QD})\right) - \left[
    \textbf{w}_\text{QD}^\top m(x\mid\boldsymbol\theta_\text{QD})
    \right]^2.
\end{align}

\subsubsection{Bayesian Query-by-Committee Acquisition Function}
The \gls{bqbc} \gls{af} is defined by \cite{riis2022bayesian}:
\begin{align}
    \alpha_\text{BQBC}(x) &:= \mathbb{V}\text{ar}_{p(\theta\mid\textbf{D}_\text{obs})}\bigg[ m(x\mid\theta)\bigg],\\
    &= \mathbb{E}_{p(\theta\mid\textbf{D}_\text{obs})}\bigg[ (m(x\mid\theta) - \hat m(x))^2\bigg].
\end{align}
The quadrature distillation approximates this without parabolic-transformation, as follows:
\begin{align}
    \alpha_\text{BQBC}(x) &\approx \textbf{w}_\text{QD}^\top \bigg[ (m(x\mid\boldsymbol\theta_\text{QD}) - \textbf{w}_\text{QD}^\top  m(x\mid\boldsymbol\theta_\text{QD}))^2\bigg].
\end{align}
A variant \gls{af}, \gls{qbmgp}, can also be approximated by the quadrature distillation without parabolic-transformation:
\begin{align}
    \alpha_\text{QB-MGP}(x) &:= \mathbb{E}_{p(\theta\mid\textbf{D}_\text{obs})}\bigg[C(x, x\mid\theta)\bigg] + \mathbb{E}_{p(\theta\mid\textbf{D}_\text{obs})}\bigg[ (m(x\mid\theta) - \hat m(x))^2\bigg].\\
    &\approx \textbf{w}_\text{QD}^\top C(x, x\mid\boldsymbol\theta_\text{QD}) + 
    \textbf{w}_\text{QD}^\top \bigg[ (m(x\mid\boldsymbol\theta_\text{QD}) - \textbf{w}_\text{QD}^\top  m(x\mid\boldsymbol\theta_\text{QD}))^2\bigg].
\end{align}

\subsection{Quadrature Distillation Algorithm} \label{app:bqweights}
\newcommand{\codecomment}[1]{\hspace{\fill}\rlap{\# #1}\phantom{Define $n-1$ test functions via Nyström}}
\begin{table}
  \caption{Quadrature distillation algorithm}
  \label{alg:qd}
  \centering
  \begin{center}
  \begin{tabular}{l}
    \toprule
    \textbf{Algorithm 2}: Quadrature distillation
    \\%
    \midrule%
    ~~1: \textbf{if} MCMC-based?\\
    ~~2: ~~~~$\Pi_\text{emp} := (\textbf{w}_\text{hyper}, \boldsymbol\theta_\text{obs}) = (\textbf{1}/M, \boldsymbol{\theta}_\text{obs}), \quad \boldsymbol\theta_\text{obs} \sim \Pi(\cdot)$ \codecomment{MCMC sampling from hyperposterior}\\
    ~~3: \textbf{elif} BQ-based?\\
    ~~4: ~~~~$\boldsymbol\theta_\text{obs} \sim \Pi^\prime(\cdot)$ \codecomment{random sampling from hyperprior} \\
    ~~5: ~~~~$\ell(\theta) \leftarrow \text{TrainGP}(\boldsymbol\theta_\text{obs}, L(\boldsymbol\theta_\text{obs}))$ \codecomment{train hyper-GP}\\%
    ~~6: ~~~~$\Pi_\text{emp} := (\textbf{w}_\text{BQ}, \boldsymbol\theta_\text{BQ}) \leftarrow \text{BayesQuad}(\ell(\cdot), \boldsymbol\theta_\text{obs}, L(\boldsymbol\theta_\text{obs})) $\codecomment{Bayesian quadrature}\\%
    ~~7: $(\textbf{w}_\text{QD}, \boldsymbol{\theta}_\text{QD}) = \text{RCHQ}(\Pi_\text{emp}, K_\text{hyper})$ \codecomment{quadrature distillation via RCHQ}\\%
    ~~8: \textbf{return }$\textbf{w}_\text{QD}, \boldsymbol\theta_\text{QD}$\\
    \bottomrule
  \end{tabular}
  \end{center}
\end{table}

The algorithm flow of the quadrature distillation is shown in Table \ref{alg:qd}. Each procedure will be explained step by step.

\section{Algorithm}
\begin{table}
  \caption{SOBER algorithm.}
  \label{alg:sober}
  \centering
  \begin{tabular}{l}
    \toprule
    \textbf{Algorithm 1}: SOBER
    \\%
    \midrule%
    ~~~~~~\textbf{Input}: prior $\pi^\prime$, hyperprior $\Pi^\prime(\theta)$,\\
    \hspace{42pt} observed dataset $\textbf{D}_\text{obs} = (\textbf{X}_\text{ob}, \textbf{y}_\text{ob})$\\
    ~~~~~~\textbf{Output}: maximum $\arg \max[\textbf{y}_\text{ob}]$, evidence $\mathbb{E}[m(x)]$\\%
    ~~1: $f \leftarrow \text{InitialiseGP}(\textbf{D}_\text{obs})$\\%
    ~~2: \textbf{while} convergence:\\%
    ~~3: ~~~~\textbf{if} FBGP:\\%
    ~~4: ~~~~~~~~$\textbf{w}_\text{QD}, \boldsymbol\theta_\text{QD} \leftarrow \text{QuadDistil}(f, \Pi^\prime(\theta))$\\%
    ~~5: ~~~~~~~~$\pi, \alpha, K(\cdot, \cdot) \leftarrow \text{FBGP}(f, \pi^\prime, \textbf{w}_\text{QD}, \boldsymbol\theta_\text{QD})$\\%
    ~~6: ~~~~\textbf{else}:\\%
    ~~7: ~~~~~~~~$\pi, \alpha, K(\cdot, \cdot) \leftarrow \text{Type-II MLE}(f)$\\%
    ~~8: ~~~~$\textbf{w}_\text{rec}, \textbf{X}_\text{rec}, \textbf{X}_\text{nys} \sim \text{Sampling}(\pi, \pi^\prime)$\\%
    ~~9: ~~~~$\textbf{X}_\text{batch}, \textbf{w}_\text{batch} \leftarrow \text{AutoKQ}(\textbf{w}_\text{rec}, \textbf{X}_\text{rec}, \textbf{X}_\text{nys}, \alpha, K(\cdot, \cdot))$\\%
    10: ~~~~$\textbf{y}_\text{batch} = \text{ParallelQuery}(f_\text{true}(\textbf{X}_\text{batch}))$\\
    11: ~~~~$\textbf{D}_\text{obs} \leftarrow \textbf{D}_\text{obs} \cup \textbf{D}_\text{batch}$\\%
    12: ~~~~$f \leftarrow \text{UpdateGP}(f, \textbf{D}_\text{obs})$\\
    13: ~~~~$\pi^\prime \leftarrow \pi$\\
    14: ~~~~$\mathbb{E}[m(x)], \mathbb{V}\text{ar}[m(x)] \leftarrow \text{KQ}(f, \textbf{X}_\text{batch}, \textbf{w}_\text{batch})$ \\
    15: \textbf{return} $\arg \max[\textbf{y}_\text{ob}], \mathbb{E}[m(x)]$\\
    \bottomrule
  \end{tabular}
\end{table}

The whole algorithm flow of \gls{sober} is shown in Table \ref{alg:sober}. QuadDistil is short for quadrature distillation explained in Section \ref{sec:qd}. Sampling procedure is deliniated in Section \ref{app:sampling}. AutoKQ is short for automatic kernel quadrature selection, explained in Section \ref{sec:akq}.

\subsection{Sampling from \texorpdfstring{$\pi$}{p}}
\gls{sober} is a sample-based gradient-free approach, and so can handle discrete, continuous or mixed inputs. The only difference is the sampler for $\textbf{X}_\text{rec}$.
The simplest scenario is if all discrete candidates are available \textit{a priori} and enumerable. As \gls{rchq} accepts weighted samples $\pi_\text{emp} =(\textbf{w}_\text{rec}, \textbf{X}_\text{rec})$ for importance sampling, all we have to do is to calculate the weights $\textbf{w}_\text{rec}$. This is simply the normalised posterior $\pi(\textbf{X}_\text{rec}) / \big[ \pi(\textbf{X}_\text{rec}) \cdot \textbf{1} \big]$. If all combinations are innumerable or unavailable, we sample $\textbf{X}_\text{rec}$ from the discrete prior $\pi^\prime$, which the user can define the arbitrarily. Once sampled, the procedure is the same: we compute $\textbf{w}_\text{rec}$, then pass the empirical measure $\pi_\text{emp}$ to \gls{rchq}. We update the hyperparameters of the prior $\pi^\prime$ via \gls{mle} from the weighted sample $(\textbf{w}_\text{rec}, \textbf{X}_\text{rec})$.
Continuous space can be regarded as innumerable discrete space, so it can be handled similarly. The only difference is the prior update. We use weighted \gls{kde} for the update, for speed and flexibility.
Mixed space is the combination of discrete and continuous space, which also can be regarded as innumerable discrete space. The prior update is the combination of the above two by assuming the discrete and continuous parameters are independent. Importantly, the prior does not need to precisely approximate $\pi$ as the importance weights $\textbf{w}_\text{rec}$ will correct the difference.

\subsection{Automatic Kernel Quadrature Selection} \label{sec:akq}
\begin{table}[h!]
  \caption{AutoKQ selection algorithm}
  \label{alg:autokq}
  \centering
  \begin{center}
  \begin{tabular}{l}
    \toprule
    \textbf{Algorithm 4}: AutoKQ selection
    \\%
    \midrule%
    ~~1: $\textbf{X}_\text{rchq}, \textbf{w}_\text{rchq}, \mathbb{V}\text{ar}[m(x)]_\text{rchq}
        \leftarrow \text{RunRCHQ}(\textbf{w}_\text{rec}, \textbf{X}_\text{rec}, \boldsymbol\varphi(\cdot), \alpha(\cdot), \textbf{X}_\text{nys}, f(\cdot))$\\
    ~~4: $\textbf{X}_\text{kt}, \textbf{w}_\text{kt}, \mathbb{V}\text{ar}[m(x)]_\text{kt} \leftarrow \text{RunKernelThinning}(\textbf{w}_\text{rec}, \textbf{X}_\text{rec}, \boldsymbol\varphi(\cdot), \alpha(\cdot), \textbf{X}_\text{nys}, f(\cdot)$\\
    ~~6: \textbf{if} $\mathbb{V}\text{ar}[m(x)]_\text{rchq} < \mathbb{V}\text{ar}[m(x)]_\text{kt}$:\\
    ~~7: ~~~~~\textbf{return} $\textbf{X}_\text{rchq}, \textbf{w}_\text{rchq}$\\
    ~~8: \textbf{else}:\\
    ~~9: ~~~~~\textbf{return} $\textbf{X}_\text{kt}, \textbf{w}_\text{kt}$\\
    \midrule
    \textbf{function} RunRCHQ($\textbf{w}_\text{rec}, \textbf{X}_\text{rec}, \boldsymbol\varphi(\cdot), \alpha(\cdot), \textbf{X}_\text{nys}, f(\cdot))$:\\
    ~~1: $\boldsymbol\varphi(\cdot)
        \leftarrow \text{Nyström}(\textbf{X}_\text{nys}, f(\cdot))$ \\%
    ~~2: $\textbf{X}_\text{rchq}, \textbf{w}_\text{rchq} \leftarrow \text{RCHQ}(\textbf{w}_\text{rec}, \textbf{X}_\text{rec}, \boldsymbol\varphi(\cdot), \alpha(\cdot))$\\%
    ~~3: $\mathbb{V}\text{ar}[m(x)]_\text{rchq} \leftarrow \text{KQ}(f(\cdot), \textbf{X}_\text{batch}, \textbf{w}_\text{batch})$\\
    ~~4: \textbf{return} $\textbf{X}_\text{rchq}, \textbf{w}_\text{rchq}, \mathbb{V}\text{ar}[m(x)]_\text{rchq}$\\
    \midrule
    \textbf{function} RunKernelThinning($\textbf{w}_\text{rec}, \textbf{X}_\text{rec}, \boldsymbol\varphi(\cdot), \alpha(\cdot), \textbf{X}_\text{nys}, f(\cdot))$:\\
    ~~4: $\textbf{X}_\text{kt}, \textbf{w}_\text{kt} \leftarrow \text{KernelThinning}(\textbf{X}_\text{rec}, f(\cdot), \alpha(\cdot))$\\%
    ~~5: $\mathbb{V}\text{ar}[m(x)]_\text{kt} \leftarrow \text{KQ}(f(\cdot), \textbf{X}_\text{batch}, \textbf{w}_\text{batch})$\\
    ~~4: \textbf{return} $\textbf{X}_\text{kt}, \textbf{w}_\text{kt}, \mathbb{V}\text{ar}[m(x)]_\text{kt}$\\
    \bottomrule
  \end{tabular}
  \end{center}
\end{table}

Table \ref{alg:autokq} illustrates the algorithm flow of automatic kernel quadrature selection algorithm. We compare the worst-case integration error of each algorithm, then pick the batch queries of which integration error is smaller. The choice between these two \gls{kq} methods can be made automatically by comparing the worst-case error $\text{wce}(Q_{\textbf{w}_\text{batch}, \textbf{X}_\text{batch}})$:
\begin{align}
\begin{split}
    &\text{wce}(Q_{\textbf{w}_\text{batch}, \textbf{X}_\text{batch}})\\
    &:= \sup_{\lVert f\rVert_{\mathcal{H}}\le 1}\lvert \textbf{w}_\text{batch} f(\textbf{X}_\text{batch}) - \int f(x)\pi(x)\text{d} x\rvert,\\
    &\approx \textbf{w}_\text{batch}^\top K(\textbf{X}_\text{batch}, \textbf{X}_\text{batch}) \textbf{w}_\text{batch} - 2 \textbf{w}_\text{batch}^\top K(\textbf{X}_\text{batch}, \textbf{X}_\text{rec}) \textbf{w}_\text{rec} + \textbf{w}_\text{rec}^\top K(\textbf{X}_\text{rec}, \textbf{X}_\text{rec}) \textbf{w}_\text{rec},
\end{split} \label{eq:wce}
\end{align}
where $\mathcal{H}$ is the reproducing kernel Hilbert space.
The third term in Eq.\ (\ref{eq:wce}) is not dependent on the \gls{kq} methods, so we can avoid expensive $N \times N$ computations. \gls{rchq} is selected in the early stage because the smooth kernel makes the eigenvalue decay short-tailed. In the late stage, the kernel thinning is chosen when the region is narrowed.

\subsection{Sampling algorithm} \label{app:sampling}
The algorithm flow of the sampling is shown in the Table \ref{alg:sample}. The details will be explained step by step.

\begin{table}[h!]
  \caption{Sampling algorithm}
  \label{alg:sample}
  \centering
  \begin{center}
  \begin{tabular}{l}
    \toprule
    \textbf{Algorithm 3}: Subsampling
    \\%
    \midrule%
    ~~1: $\textbf{X}_\text{rec} \sim \pi^\prime(\cdot)$ \codecomment{sampling from prior}\\%
    ~~2: $\textbf{w}_\text{rec} = \frac{L(\textbf{X}_\text{rec})}{\pi^\prime(\textbf{X}_\text{rec})} \cdot \frac{\pi^\prime(\textbf{X}_\text{rec})^\top \textbf{1}}{L(\textbf{X}_\text{rec})^\top \textbf{1}}$ \codecomment{compute the weights}\\%
    ~~3: \textbf{if} len$(\textbf{w}_\text{rec} > 0) < n$ :\\%
    ~~4: ~~~~~$\pi^\prime(\cdot) \leftarrow \pi^\prime_\text{initial}(\cdot)$ \codecomment{return to the initial prior when overexploitive}\\%
    ~~5: \textbf{if} continuous:\\%
    ~~6: ~~~~~$\pi(\cdot) = \text{WKDE}(\textbf{w}_\text{rec}, \textbf{X}_\text{rec})$ \codecomment{weighted kernel density estimation}\\%
    ~~7: ~~~~~$\textbf{X}_\text{rec} \sim \pi(\cdot)$ \codecomment{resample from WKDE}\\%
    ~~8: ~~~~~$\textbf{w}_\text{rec} = \frac{L(\textbf{X}_\text{rec})}{\pi(\textbf{X}_\text{rec})} \cdot \frac{\pi(\textbf{X}_\text{rec})^\top \textbf{1}}{L(\textbf{X}_\text{rec})^\top \textbf{1}}$ \codecomment{recompute the weights}\\%
    ~~9: \textbf{else if} discrete and enumerable:\\%
    10: ~~~~~$\textbf{X}_\text{rec} = \pi^\prime(\cdot)$ \codecomment{all discrete candidates}\\%
    11: ~~~~~$\textbf{w}_\text{rec} = \frac{L(\textbf{X}_\text{rec})}{L(\textbf{X}_\text{rec})^\top \textbf{1}}$ \codecomment{normalised weights}\\%
    12: \textbf{else if} innumerable discrete:\\%
    13: ~~~~~$\pi(\cdot) \leftarrow \text{OptHypersMLE}(\pi^\prime(\cdot), \textbf{w}_\text{rec}, \textbf{X}_\text{rec})$ \codecomment{MLE hyperparameter optimisation}\\%
    14: ~~~~~$\textbf{X}_\text{rec} \sim \pi(\cdot)$ \codecomment{resample from WKDE}\\%
    15: ~~~~~$\textbf{w}_\text{rec} = \frac{L(\textbf{X}_\text{rec})}{\pi(\textbf{X}_\text{rec})} \cdot \frac{\pi(\textbf{X}_\text{rec})^\top \textbf{1}}{L(\textbf{X}_\text{rec})^\top \textbf{1}}$ \codecomment{recompute the weights}\\%
    16: \textbf{else} mixed:\\%
    17: ~~~~~$\pi(\cdot) \leftarrow \text{CombineBothPrior}(\pi^\prime(\cdot), \textbf{w}_\text{rec}, \textbf{X}_\text{rec})$ \codecomment{Combine continuous and discrete prior}\\%
    18: ~~~~~$\textbf{w}_\text{rec} = \frac{L(\textbf{X}_\text{rec})}{\pi(\textbf{X}_\text{rec})} \cdot \frac{\pi(\textbf{X}_\text{rec})^\top \textbf{1}}{L(\textbf{X}_\text{rec})^\top \textbf{1}}$ \codecomment{recompute the weights}\\%
    19: $\textbf{X}_\text{nys} \sim \text{Deweighted}(\textbf{w}_\text{rec}, \textbf{X}_\text{rec})$ \codecomment{deweighted random subset extraction}\\%
    20: \textbf{return} $\textbf{w}_\text{rec}, \textbf{X}_\text{rec}, \textbf{X}_\text{nys}$\\
    \bottomrule
  \end{tabular}
  \end{center}
\end{table}

\subsubsection{Weighted Kernel Density Estimation}
The mean and covariance of the \gls{wkde} is estimated with the unbiased data covariance matrix given by:
\begin{align}
    \mu_\text{wkde} &:= \textbf{w}_\text{rec}^\top \textbf{X}_\text{rec},\\
    \boldsymbol\Sigma_\text{wkde} &:= \frac{\textbf{w}_\text{rec}^\top \textbf{1}}{(\textbf{w}_\text{rec}^\top \textbf{1})^2 - (\textbf{w}_\text{rec}^2)^\top \textbf{1}} \sum_i^N w_\text{i, rec} (X_\text{i, rec} - \mu_\text{wkde})^T (X_\text{i, rec} - \mu_\text{wkde}),\\
    &:= \frac{1}{1 - (\textbf{w}_\text{rec}^2)^\top \textbf{1}} \sum_i^N w_\text{i, rec} (X_\text{i, rec} - \mu_\text{wkde})^T (X_\text{i, rec} - \mu_\text{wkde}),
\end{align}
where $X_\text{i, rec} \in \textbf{X}_\text{rec}$ and  $w_\text{i, rec} \in \textbf{w}_\text{rec}$ is the $i$-th element of $\textbf{X}_\text{rec}$ and $\textbf{w}_\text{rec}$, respectively. The bandwidth of the kernel is estimated by the Scott's method \cite{terrell1992variable}. 

\subsubsection{Maximum Likelihood Estimation of Discrete Prior}
The optimisation of hyperparamters of the discrete prior distributions was done via \gls{mle} from the weighted samples $(\textbf{w}_\text{rec}, \textbf{X}_\text{rec})$.
We denote the \gls{pdf} of Bernoulli distribution (binary) and the categorical distribution as $\text{Bernoul}(x; \textbf{w}_\text{Ber}), \text{Categor}(x; \textbf{w}_\text{Cat})$, where $\textbf{w}_\text{Ber} \in \mathbb{R}^{d}$ and $\textbf{w}_\text{Cat} \in \mathbb{R}^{d \times C}$ are the weights hyperparaeters, $C$ is the number of categories in the input parameters.
The weighted log-\gls{pdf} can be expressed as follows:
\begin{align}
    \text{LL} := \textbf{w}_\text{rec}^\top \log \text{Bernoul}(\textbf{X}_\text{rec}; \textbf{w}_\text{Ber})\\
    \text{LL} := \textbf{w}_\text{rec}^\top \log \text{Categor}(\textbf{X}_\text{rec}; \textbf{w}_\text{Cat})
\end{align}
We optimise each weight hyperparameters via maximising the log-likelihood (LL) via L-BFGS-B \cite{liu1989limited}. PyTorch \cite{paszke2019pytorch} auto-differentiation gives the gradient for L-BFGS-B. To make weights bounded $[0,1]$, we transformed original LL space via the sigmoid function during optimisation. We start the optimisation of weights to be equal chance (discrete uniform distribution).

\subsubsection{Deweighted sampling}
Samples for the Nystr{\"o}m method are better to be spatially sparse to well represent the whole kernel shape. We adopt the deweighted sampling to constract the small subset of uniformly distributed samples $\textbf{X}_\text{nys}$ from the weighted samples $(\textbf{w}_\text{rec}, \textbf{X}_\text{rec})$. We resample from the categorical distribution with the inverse weights $(1/\textbf{w}_\text{rec})$, then the resampled samples are uniformly distributed.

\section{Simulation-based inference}\label{app:lfi}
\subsection{Simulation-based inference}
The simulator emulates typically time-evolving signals from the physical device modelled by simultaneous differential equations. The solution of the differential equation is basically not analytical, requiring numerical approximation such as the finite element method. Each equation has parameters, such as coefficients of differential terms, which determine the signal shape. Estimating the parameters that can reproduce the observed signal is a typically tricky task because simulation is not differentiable with regard to each parameter. Although auto-differentiation can mitigate this problem, the parameter posterior is typically multimodal, so local optimisation algorithms based on differentiation struggles to find the global optimum. More importantly, this inverse problem often has no unique solution mathematically. Hence, rather than estimating one deterministic parameter set, inferring the parameter posterior is more practically important. Moreover, having dozens of plausible simulators with differing levels of assumption is a common situation where we need to select the parsimonious model that best describes the given dataset. Bayesian model evidence can provide a selection criterion. Therefore, estimating both Bayesian model evidence \textit{and} parameter posterior is a frequent desideratum in practice. Furthermore, running simulators is expensive to evaluate, so parallelising the computation via computer clusters is of practical importance.

Let $\textbf{y}_\text{obs}$ be the observed signal from the physical device, and we wish to estimate the simulator parameters $\Theta$. This can be formulated as Bayesian inference, given by:
\begin{align}
p(\Theta) &:= \pi^\prime(\Theta) := \mathcal{N}(\Theta; \mu_\pi, \boldsymbol\Sigma_\pi)\\
p(\textbf{D}_\text{obs} | \Theta, M) &:= \ell_\text{true}(\Theta) := \prod_j^m \mathcal{N}(\text{err}_j(\theta); \textbf{0}, \sigma^2_\text{noise}),\\
p(\textbf{D}_\text{obs} | M) &:= 
\mathcal{N} 
\left( \mathop{\mathbb{E}}_{x\in\pi}[\ell_\text{true}(\Theta)],     \mathop{\mathbb{V}\text{ar}}_{x\in\pi}[\ell_\text{true}(\Theta)] 
\right),\\
p(\Theta | \textbf{D}_\text{obs}, M) &= \frac{p(\textbf{D}_\text{obs} | \Theta, M) p(\Theta)}{p(\textbf{D}_\text{obs} | M)} = \frac{\ell_\text{true}(\Theta)\pi(\Theta)}{\mathop{\mathbb{E}}_{x\in\pi}[\ell_\text{true}(\Theta)]}, \label{eq:posterior}
\end{align}
where
\begin{align}
\textbf{D}_\text{obs} &:= \{\textbf{x}_\text{obs},  \textbf{y}_\text{obs} \} \in \mathbb{R}^{m \times 1},\\    \theta &:= \{ \theta_i\} \in \mathbb{R}^{d-1},\\ \Theta &:= \{ \theta, \sigma^2_\text{noise} \} \in \mathbb{R}^{d},\\    y_{\text{sim}, j}(\theta) &:= M(\theta, \textbf{x}_\text{obs}),\\ \text{err}_j(\theta) &:= \left[y_{\text{obs}, j} - y_{\text{sim}, j}(\theta) \right]^2.
\end{align}

$M(\theta, \textbf{x}_\text{obs})$ is the simulation model, which returns the prediction $y_{\text{sim}, j}(\theta)$ at given simulation parameter $\theta$ at the $j$-th time step. We wish to estimate the model evidence $\mathop{\mathbb{E}}_{x\in\pi}[\ell_\text{true}(\Theta)]$ and the parameter posterior $p(\Theta | \textbf{D}_\text{obs}, M)$.

\subsection{Bayesian Quadrature Formulation} \label{app:bqmodelling}
In naive \gls{bq}, we place \gls{gp} on the likelihood as such:
\begin{align}
\ell(\Theta) &\sim \mathcal{GP}\big(
\ell(\Theta); \mu_\ell(\Theta), \sigma_\ell(\Theta, \Theta^\prime)
\big).
\end{align}
The evidence can be estimated with \gls{basq} formulation, given by:
\begin{subequations}
\begin{align}
    \mathop{\mathbb{E}}_{x \in \pi} [ m(x) ] &\approx \textbf{w}_\text{batch}^\top m(\textbf{X}_\text{batch}), \label{eq:evidenceRCHQ}\\
    \begin{split}
    \mathop{\mathbb{V}\text{ar}}_{x \in \pi} [ m(x)] &\approx \textbf{w}_\text{batch}^\top C(\textbf{X}_\text{batch}, \textbf{X}_\text{batch}) \textbf{w}_\text{batch} - 2 \textbf{w}_\text{batch}^\top C(\textbf{X}_\text{batch}, \textbf{X}_\text{rec}) \textbf{w}_\text{rec} + \textbf{w}_\text{rec}^\top C(\textbf{X}_\text{rec}, \textbf{X}_\text{rec}) \textbf{w}_\text{rec}.
    \end{split}
\end{align}
\end{subequations}
The posterior can be estimated with the surrogate model and the estimated evidence via Eq.\ (\ref{eq:posterior}).

However, the likelihood is typically transformed into the logarithmic space because its dynamic range is wider than the numerical over-/underflow limits. Thus, log-warped \gls{gp} \cite{osborne2012active, chai2019improving, adachi2022bayesian} is often applied. Particularly, we consider moment-matched log-transformed (MMLT) \cite{chai2019improving} \gls{gp}, modelled as such:
\begin{subequations}
\begin{align}
    &f(x) = \exp [g(x)] - 1,\\
    &\quad\quad := \mathcal{GP}\big( f(x); m(x), C(x,x) \big), \\
    &g(x) := \mathcal{GP}\big( g(\cdot); m_g(\cdot), C_g(\cdot,\cdot) \big),\\
    &m(x) := \exp \left[m_g(x) + \frac{1}{2}C_g(x,x)\right],\\
    &C(x,x^\prime) := m_g(x)m_g(x^\prime) \left[C_g(x,x) - 1\right]. \label{eq:mmltkernel}
\end{align}
\end{subequations}

The warped \gls{gp} stores log-transformed values $\textbf{y}_g = \log(\textbf{y} + 1)$, so we can avoid the over-/underflows. Adachi et al.\ \cite{adachi2022bayesian} further extended MMLT \gls{gp} so as to accommodate with \gls{basq} modelling. They adopted the four-layered \gls{gp} combining MMLT and parabolic transformation. The reason why they add the parabolic transformation is to copy the exponetiated function information to not only the surrogate function but also prior update. However, this deep warped structure causes additional predictive errors due to the cumulative approximation errors from each layer's moment-matching method.

\subsection{Likelihood-free Inference Formulation}
Alternately, \gls{bo}-based \gls{lfi} \cite{gutmann2016bayesian} models \gls{gp} differently. They placed \gls{gp} on the discrepancy, rather than the likelihood, defined as :
\begin{align}
\Delta_\text{true}(\theta) &:= \log || \textbf{y}_\text{obs} - \textbf{y}_\text{sim}(\theta) ||\\
\Delta(\Theta) &\sim \mathcal{GP}\big(
\Delta(\Theta); \mu_\Delta(\theta), \sigma_\Delta(\theta, \theta^\prime)
\big)
\end{align}
\gls{lfi} adopts the tentative likelihood defined by Eq.\ \ref{eq:templik} as the likelihood at each iteration. The true likelihood can be estimated \textit{a posteriori}.
The benefits of this modelling are as follows:
\begin{compactenum}
    \item Avoiding extreme dynamic range of likelihood; $\Delta_\text{true}(\theta)$ has much more moderate range.
    \item We can reformulate BQ as BO. BO is more suitable for solving simulation-based inference as only the vicinity of the MAP location has meaningful value. As almost everywhere has zero likelihood, so BQ formulation is over-exploring if the prior is misspecified.
    \item We can obtain the “temporary” likelihood $L(\theta)$ that approaches the true likelihood $\ell_\text{true}(\Theta)$ asymptotically over iterations. This likelihood can be regarded as "updated prior". This can also mitigate the prior misspecification.
\end{compactenum}
They reformulate the posterior inference as the \gls{bo} to find the global minimum of the discrepancy $\Delta_\text{true}(\theta)$. The resulting \gls{gp} surrogate model is used to approximate the posterior. They do not go beyond the posterior inference, so evidence estimation cannot be done with BOLFI.

\subsection{SOBER-LFI Formulation}
We wish to take the best of both world; \gls{lfi} \gls{gp} modelling suitable for sampling and exact evidence estimation via \gls{bq} modelling. Thus, we adopt the dual \glspl{gp}; one for sampling, and the other for \gls{bq} modelling. While the sampling \gls{gp} is modelled with the inverse discrepancy ($-\Delta_\text{true}(\theta)$ so as to be the maximisation objective), the \gls{bq} \gls{gp} is modelled with log-likelihood with MMLT \gls{gp}. Importantly, we can query both $\{\Delta_\text{true}(\theta), \ell_\text{true}(\Theta)\}$ with negligible overhead as the time-consuming part is $\textbf{y}_\text{sim}(\theta)$. Once we get $\textbf{y}_\text{sim}(\theta)$, calculating both $\{\Delta_\text{true}(\theta), \ell_\text{true}(\Theta)\}$ are very cheap.

The sampling \gls{gp} is used for setting up the sampling function $\pi$, in the same manner explained in Section \ref{app:sampling}. One difference is that the $\pi$ becomes extremely sharper than the \gls{bo} task. \gls{wkde}-based sampling can fail to sampling from $\pi$. Hence, we adopted elliptical slice sampling (ESS) \cite{murray2010elliptical}. Importance sampling permits using all of the samples from ESS without the burn-in period. The weights can be calculated via the $\pi$ defined with the sampling \gls{gp}. Note that ESS is more expensive than \gls{wkde}, so the additional overhead had to be produced instead. As such, the sampling \gls{gp} constructs the empirical measure $\pi_\text{emp} = (\textbf{w}_\text{rec}, \textbf{X}_\text{rec} )$.

On the other hand, \gls{bq} \gls{gp} constructs the surrogate model for likelihood. The posterior and evidence inference can be made in the same manner explained in Section \ref{app:bqmodelling}. 

Batch acquisition via objective \gls{rchq} becomes a mix of both \glspl{gp}. The kernel is defined by the \gls{bq} \gls{gp} in Eq.\ \ref{eq:mmltkernel}. The objective with \gls{af} is defined by the sampling \gls{gp}.

\section{Experiments}
\subsection{Batch Bayesian Optimisation}\label{app:expBO}
\begin{figure*}
  \centering
  \includegraphics[width=1.00\textwidth]{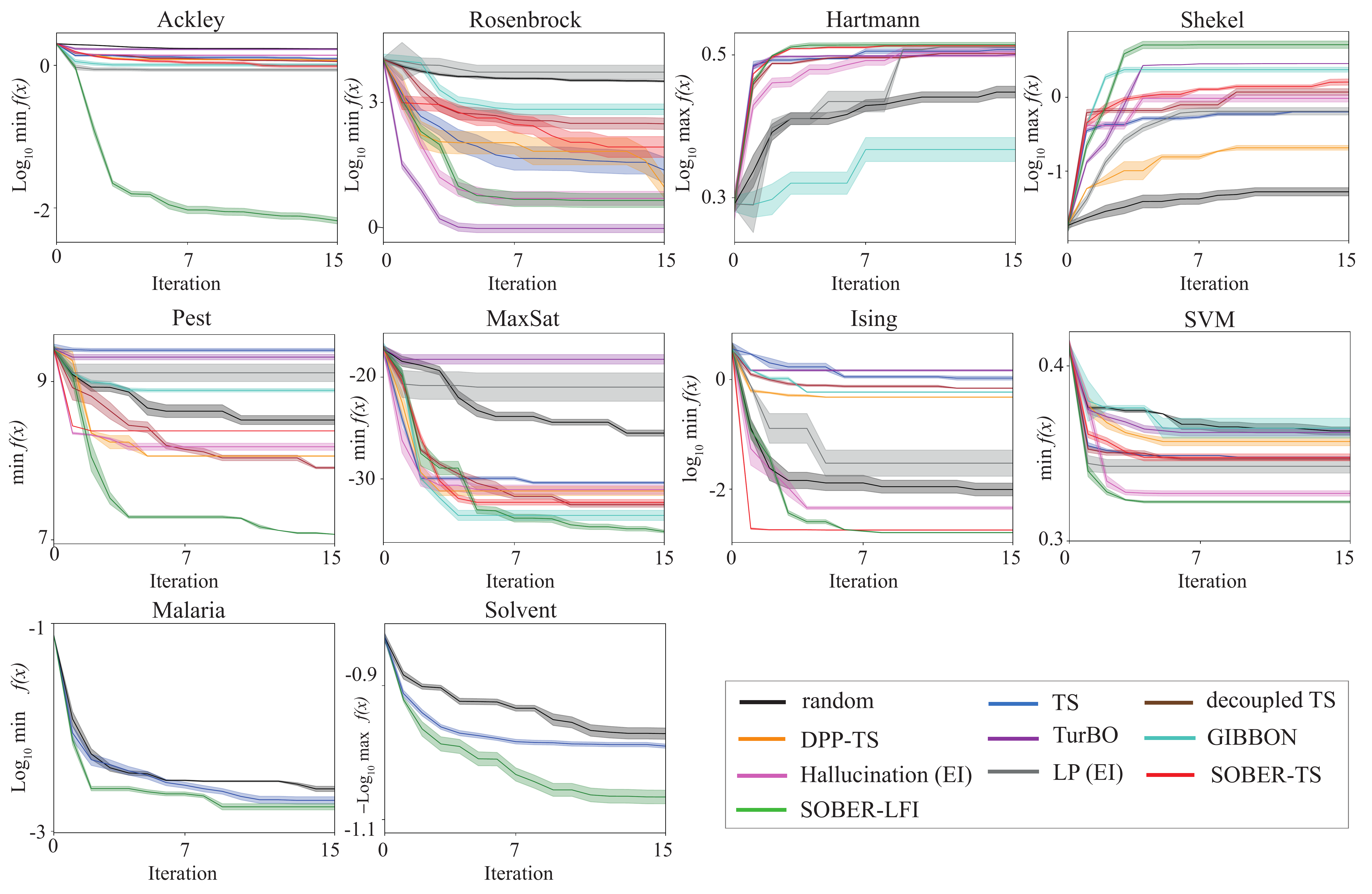}
  \caption{We evaluate \gls{sober} across 4 synthetic functions and 6 real-world drug discovery datasets. Top: Log regret or log best observations, Bottom: Log overhead in seconds as function of iterations. Lines and shaded area denote mean $\pm$ 1 standard error. The batch size is 100 or 200 (see Table \ref{tab:exp}).}
  \label{fig:expbo}
\end{figure*}
We examined our method, \gls{sober}, with the following 10 datasets. All experiments are averaged over 16 iterations with varied random seeds.
\textbf{random} is the random samples drawn from the prior distribution for each task. As these random samples are aware of categorical or mixed variables, it often performs better than baseline methods that cannot handle them (e.g. TurBO, LP). \textbf{TS} and \textbf{decoupled TS} was implemented with BoTorch library \cite{balandat2020botorch}. The candidates are sampled from the prior distribution, and each \gls{ts} algorithm took the argmax of function samples over candidates.
\textbf{DPP-TS} is provided by the author of the paper \cite{nava2022diversified}. \textbf{TurBO} and \textbf{GIBBON} were implemented based on the official tutorials of BoTorch. Both \textbf{Hallucination} and \textbf{LP} were implemented using GPyTorch \cite{gardner2018gpytorch}, and we select the standard \gls{ei} \gls{af}. Both \gls{sober}-\gls{ts} and \gls{sober}-\gls{lfi} experiments were conducted without \gls{fbgp}, and the hyperparameters $(N, M)$ are fixed with $(20,000, 500)$ throughout the experiments. The \gls{sober} software is coded with GPyTorch, BoTorch, and \gls{basq} code \cite{adachi2022fast}. All codes are provided in the supplementary and will be open-sourced.

\paragraph{Synthetic: Ackley function}
Ackley funciton is defined as:
\begin{align}
f(x) := - a \exp \left[ -b \sqrt{\frac{1}{d} \sum_{i=1}^d x_i^2} \right] - \exp \left[
\frac{1}{d} \sum_{i=1}^d \cos (c x_i) \right] + a + \exp(1)
\end{align}
where $a = 20, c = 2\pi, d = 23$. We take the negative Ackley function as the objective of \gls{bo} to make this optimisation problem maximisation. We modified the original Ackley function into a 23-dimensional function with the mixed spaces of 3 continuous and 20 binary inputs from $[0, 1]^{20}$, following \cite{daulton2022bayesian}. The batch size $n$ is 200. The continuous prior is the uniform distribution ranging from [-1, 1]. The binary prior is the Bernoulli distribution with unbiased weights 0.5. We assume each of continuous and binary priors at each dimension are independent.

\paragraph{Synthetic: Rosenbrock function}
Rosenbrock function is defined as:
\begin{align}
f(x) := \left[
\sum_{i=1}^{d - 1} \left\{ 100 (x_{i+1} - x_i^2)^2 + (x_i - 1)^2 \right\}
\right]
\end{align}
where $d=7$. We take the negative Rosenbrock function as the objective of \gls{bo} to make this optimisation problem maximisation. We modified the original Rosenbrock function into a 7-dimensional function with the mixed spaces of 1 continuous and 6 discrete variables, following \cite{daulton2022bayesian}. The first 1 dimension is continuous with bounds $[-4, 11]^1$. The other 6 dimensions are discretised to be categorical variables, with 4 possible values $x_1 \in \{-4, 1, 6, 11\}$. The batch size $n$ is 100. The continuous prior is the uniform distribution ranging from [-5, 10]. The discrete prior is the categorical distribution with unbiased weights 0.5. We assume each of the continuous and discrete priors at each dimension is independent.

\paragraph{Synthetic: Hartmann function}
Hartmann 6-dimensional function \cite{simulationlib} is defined as:
\begin{align}
f(x) &:= -\sum_{i=1}^4 \alpha_i \exp \left( - \sum_{j=1}^6 A_{ij} (x_j - P_{ij})^2 \right),\\
\alpha &= (1.0, 1.2, 3.0, 3.2)^\top, \\
\textbf{A} &= \begin{pmatrix}
10 & 3 & 17 & 3.5 & 1.7 & 8\\
0.05 & 10 & 17 & 0.1 & 8 & 14\\
3 & 3.5 & 1.7 & 10 & 17 & 8\\
17 & 8 & 0.05 & 10 & 0.1 & 14\\
\end{pmatrix}, \\
\textbf{P} &= \begin{pmatrix}
1312 & 1696 & 5569 & 124 & 8283 & 5886\\
2329 & 4135 & 8307 & 3736 & 1004 & 9991\\
2348 & 1451 & 3522 & 2883 & 3047 & 6650\\
4047 & 8828 & 8732 & 5743 & 1091 & 381\\
\end{pmatrix}
\end{align}
We take the negative Hartmann function as the objective of \gls{bo} to make this optimisation problem maximisation. All input variables are continuous with bounds $[0, 1]^6$. The batch size $n$ is 100. The continuous prior is the uniform distribution ranging from [0, 1].

\paragraph{Synthetic: Shekel function}
Shekel function \cite{simulationlib} is defined as:
\begin{align}
f(x) &:= -\sum_{i=1}^{10} \left( \sum_{j=1}^4 (x_j - C_{ji})^2 + \beta_i \right),\\
\beta &= \frac{1}{10} (1, 2, 2, 4, 4, 6, 3, 7, 5, 5)^\top, \\
\textbf{C} &= \begin{pmatrix}
4.0 & 1.0 & 8.0 & 6.0 & 3.0 & 2.0 & 5.0 & 8.0 & 6.0 & 7.0\\
4.0 & 1.0 & 8.0 & 6.0 & 7.0 & 9.0 & 3.0 & 1.0 & 2.0 & 3.6\\
4.0 & 1.0 & 8.0 & 6.0 & 3.0 & 2.0 & 5.0 & 8.0 & 6.0 & 7.0\\
4.0 & 1.0 & 8.0 & 6.0 & 7.0 & 9.0 & 3.0 & 1.0 & 2.0 & 3.6\\
\end{pmatrix},
\end{align}
We take the negative Shekel function as the objective of \gls{bo} to make this optimisation problem maximisation. All input variables are continuous with bounds $[0, 10]^6$. The batch size $n$ is 100. The continuous prior is the uniform distribution ranging from [0, 10]. 

\paragraph{Real-world: Pest Control}
Pest Control (Pest in the main) is proposed in \cite{oh2019combinatorial}, which is a multi-categorical optimisation problem (15 dimensions, 5 categories for each dimension). We wish to optimise the effectiveness of pesticide by choosing the 5 actions (selection of pesticides from 4 different firms, or not using any of it), but penalised by their prices. This choice is a sequential decision of 15 stages, and the objective function is expressed as the cumulative loss function with the total of both cost and the portion having pest. The batch size $n$ is 200. We set the categorical prior with equal weights for each choice (discrete uniform distribution). Code is used in \href{https://github.com/xingchenwan/Casmopolitan}{https://github.com/xingchenwan/Casmopolitan} \cite{wan2021think}.

\paragraph{Real-world: Maximum Satisifiability}
Maximum satisfiability (MaxSat in the main) is proposed in \cite{oh2019combinatorial}, which is 28 dimensional binary optimisation problem. The objective is to find boolean values that maximise the combined weighted satisfied clauses for the dataset provided by Maximum Satisfiability competition 2018. The batch size $n$ is 200. We set the Bernoulli distribution prior with equal weights (discrete uniform distribution). Both code and dataset are used in \href{https://github.com/xingchenwan/Casmopolitan}{https://github.com/xingchenwan/Casmopolitan} \cite{wan2021think}.

\paragraph{Real-world: Ising Model Sparsification}
Ising Model Sparsification (Ising in the main) is proposed in \cite{oh2019combinatorial}, which is 24 dimensional binary optimisation problem. The objective is to sparsify an Ising model using the regularised Kullback-Leibler divergence between a zero-field Ising model and the partition function, considering $4 \times 4$ grid of spins with regularisation coefficient $\lambda = 10^{-4}$. The batch size $n$ is 100. We set the Bernoulli distribution prior with equal weights (discrete uniform distribution). Code is used in \href{https://github.com/QUVA-Lab/COMBO}{https://github.com/QUVA-Lab/COMBO} \cite{oh2019combinatorial}.

\paragraph{Real-world: Support Vector Machine Feature Selection}
Support vector machine feature selection (SVM in the main) is proposed in \cite{daulton2022bayesian},  which is 23 dimensional mixed-type input optimisation problem (20 dimensional binary and 3 dimensional continuous variables). The objective is jointly performing feature selection (20 features) and hyperparameter optimisation (3 hyperparameters) for a support vector machine (SVM) trained in the CTSlice UCI dataset \cite{graf20112d, Dua2019}. The batch size $n$ is 100. We set the Bernoulli distribution prior with equal weights (discrete uniform distribution) for 20 binary inputs, and uniform prior with bounds $[0, 1]^6$. 
Code is used in \href{https://github.com/facebookresearch/bo_pr}{\text{https://github.com/facebookresearch/bo\_pr}} 
\cite{daulton2022bayesian}.

\paragraph{Real-world:  Anti-Malarial drug discovery}
\begin{figure*}
  \centering
  \includegraphics[width=0.7\textwidth]{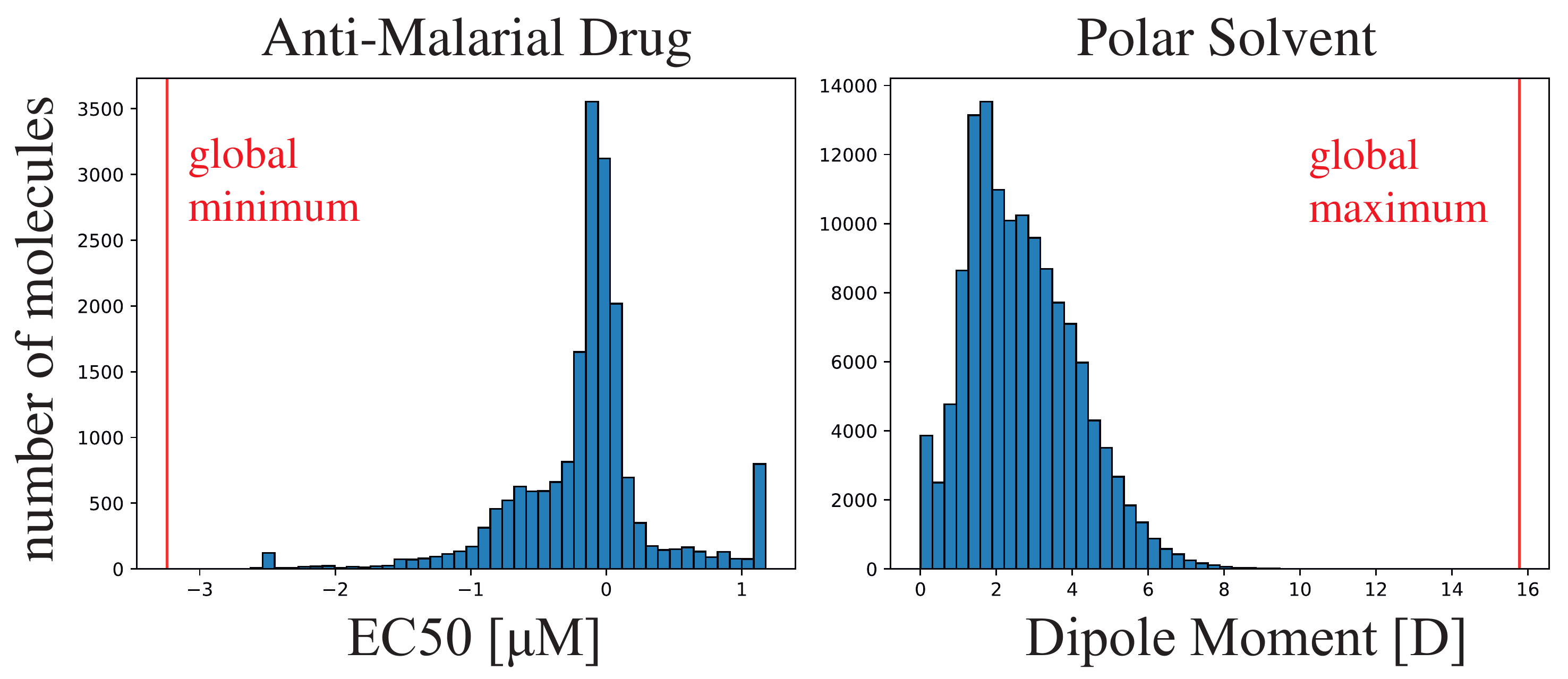}
  \caption{The histograms of the target values in the real-world drug discovery datasets}
  \label{fig:drugs}
\end{figure*}
The dataset with 20,746 small molecules represented as 2048-dimensional binary features were taken from the P. falciparum whole cell screening derived by the Novatis-GNF Malaria Box \cite{spangenberg2013open}. The target variable is the EC50 value, which is defined as the concentration of the drug which gives half the maximal response. The lower the concentration, the more effective (better) the drug. We take the nagative EC50 to make this optimisation problem maximisation. The batch size $n$ is 100. We set the categorical prior with equal weights for each molecule (discrete uniform distribution). The dataset is downloaded from \href{https://www.mmv.org/mmv-open/malaria-box/malaria-box-supporting-information}{https://www.mmv.org/mmv-open/malaria-box/malaria-box-supporting-information}. The raw molecule inputs as SMILES are converted into 2048-dimensional binary features with Gauche \href{https://github.com/leojklarner/gauche}{https://github.com/leojklarner/gauche} \cite{griffiths2022gauche}. The Tanimoto kernel is also coded with Gauche. We use the posterior predictive covariance of the \gls{gp} with Tanimoto kernel that is weighted with its predictive mean, similarly to the \cite{oh2022batch}. This drug discovery dataset is challenging for most baseline methods: For decoupled \gls{ts} in BoTorch, which is based on random Fourier features \cite{rahimi2007random}, is not compatible with the Tanimoto kernel in GAUCHE \cite{griffiths2022gauche}. \gls{sober}-\gls{ts} is also based on decoupled \gls{ts}, so it is not applicable neither. Tanimoto kernel is more computationally demanding than RBF, so \gls{dpp}-\gls{ts} and hallucination could not be finished the computation within one week. This is practically too slow, so we judged these two are not applicable for active learning purposes. TurBO is also not applicable because the Tanimoto kernel has no lengthscale hyperparameter, so TurBO cannot update the trust region. GIBBON suffers from combinatorial explosion, as enumerating all combination of both batch and drug candidates becomes infeasible $(^{20746} \text{P}_{100} \approx 3.9 \times 10^{431}$. For \gls{lp}, which explicitly assumes continuity over Euclidean space, so we cannot apply it to non-Euclidean and discrete space.

\paragraph{Real-world: Polar solvent for batteries} The dataset with 133,055 small molecules represented as 2048-dimensional features were optimised and predicted by the quantum-chemical computations using density functional theory, known as QM9 dataset \cite{ramakrishnan2014quantum}. The target variable is the dipole moment, which is basically correlated with the solvation capability in electrolytes in lithium-ion batteries, increasing the ratio of electro-mobile Li-ions. The higher the dipole moment becomes, the larger (better) the ionic conductivity does. The batch size $n$ is 200.  We set the categorical prior with equal weights for each molecule (discrete uniform distribution). The dataset is downloaded from \href{http://quantum-machine.org/datasets/}{http://quantum-machine.org/datasets/} \cite{ramakrishnan2014quantum}. The coding was done with Gauche \cite{griffiths2022gauche}. Most baseline methods cannot be applied with the same reasons explained above.

Figure \ref{fig:drugs} shows the distribution of target values in two real-world drug discovery datasets. The optimal molecules are outliers from the dataset distribution, so it clearly shows these tasks are needle-in-the-haystack situations.

\subsection{Batch Bayesian Quadrature}\label{app:expBQ}
We tested our algorithm, \gls{sober}, with the simulation-based inference tasks as the batch \gls{bq} method. All experiments are averaged over 16 iterations with varied random seeds. The number of candidate samples drawn from the prior distribution is fixed to be 20,000 $(N=20,000)$ for a fair comparison. As the ground truth of posterior and evidence cannot be obtained for the simulation-based inference, we use the empirical metric to evaluate the quality of each inference. For posterior evaluation, we drew the 10,000 test samples from the normal distribution centered at the ground truth parameters and the covariance with the identity matrix of which each element is $5 \times 10^{-6}$. Then, we computed the root-mean-squared error (RMSE) between the estimated log-likelihood and true log-likelihood. For evidence, we simply adopted the negative of estimated evidence. We use the truncated multivariate normal distribution (TMVN) as the prior. Efficient random sampling from TMVN is coded with \cite{botev2017normal}. For computing the probability density function of TMVN, we compute the normalising constant of TMVN (cumulative density function of multivariate normal distribution) using \cite{marmin2015differentiating}.

\paragraph{Real-world: 2 RC Pairs ECM} 2 RC Pair equivalent circuit model (ECM) is the simplest lithium-ion battery simulator with 6-dimensional continuous variables \cite{adachi2022bayesian}. We generated synthetic signal using the model with 100 frequency steps equispaced over log-angular frequency regime, then added the Gaussian noise with the amplitude of $\exp(1)$ to the $R_\text{total} = \exp(2)$ signal from the canonical ECM.

\paragraph{Real-world: 5 RC Pairs ECM} 5 RC Pair ECM is more complex lithium-ion battery simulator with 12-dimensional continuous variables \cite{adachi2022bayesian}. We generated synthetic signal using the model with 100 frequency steps equispaced over log-angular frequency regime, then added the Gaussian noise with the amplitude of $\exp(1)$ to the $R_\text{total} = \exp(2)$ signal from the canonical ECM.

\subsection{Additional Experiments}\label{app:addexp}
\subsubsection{Visualising Multiple Global Maxima Case}
\begin{figure}
  \centering
  \includegraphics[width=1\textwidth]{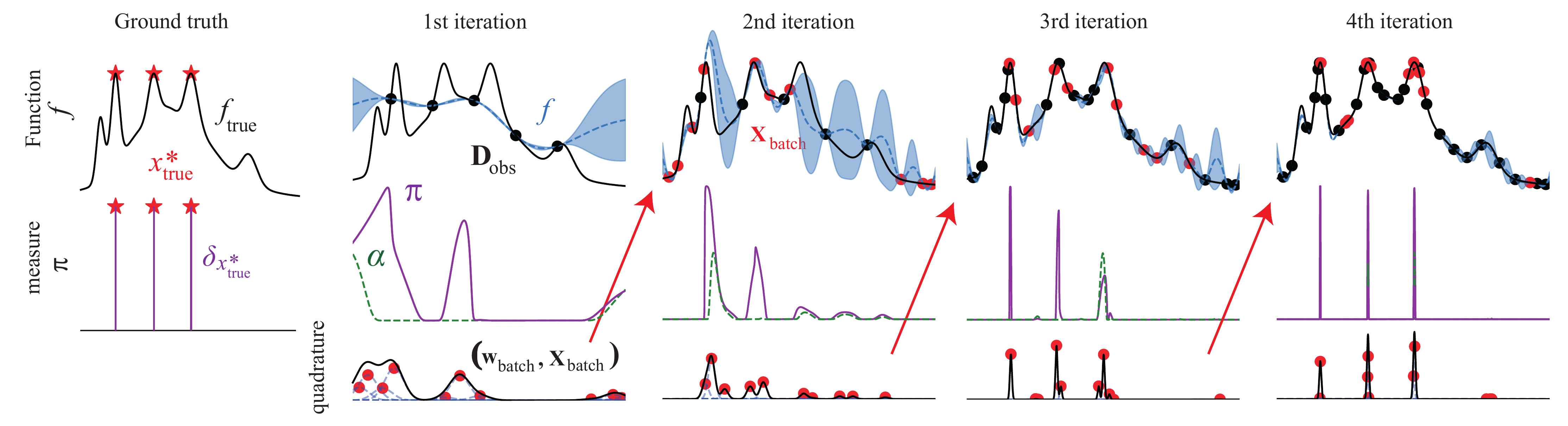}
  \caption{Visualising the multiple true global maxima case}
  \label{fig:mplgt}
\end{figure}
We visualised the case of multiple true global maxima. As shown in Figure \ref{fig:mplgt}, $\pi$ converged to multiple delta distributions.

\subsubsection{Empirical Convergence Analysis}
\begin{figure}
  \centering
  \includegraphics[width=1\textwidth]{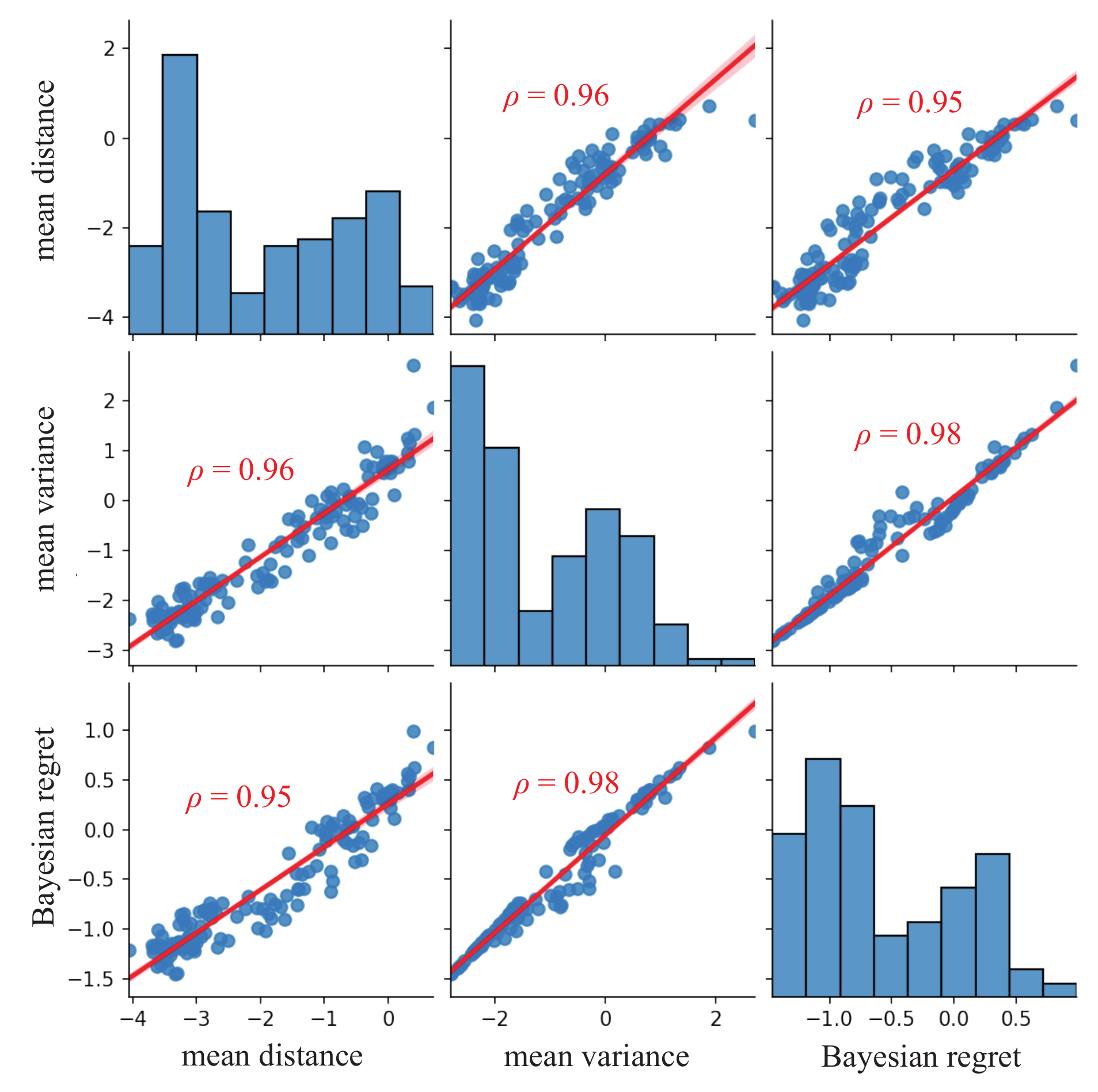}
  \caption{The correlations between regret and measure optimisation}
  \label{fig:corr}
\end{figure}
We empirically investigated the relationship between regret and $\pi$ shrinkage. Recall that empirical measure $\pi_\text{emp} = (\textbf{w}_\text{rec}, \textbf{X}_\text{rec})$ is sampled from $\pi$, and batch measure $\pi_\text{batch} = (\textbf{w}_\text{batch}, \textbf{X}_\text{batch})$ is the subset further extracted from $\pi_\text{emp}$. 
As such, all measures approximate the same distribution $\pi$, only the level of discretisation differs.

Given two measures $\pi_\text{emp}$ and $\pi_\text{batch}$, we consider the following two metrics for $\pi$ shrinkage; \gls{mv} $\mathbb{V}\text{ar}[\pi(x)]$ and \gls{md} $|x^*_\text{true} - \mathbb{E}[\pi(x)]|$. 
The empirical measure $\pi_\text{emp}$ can approximate these as such:
\begin{subequations}
    \begin{align}
        \mathbb{E}[\pi(x)] &:= \int x \text{d} \pi(x) \approx \textbf{w}_\text{rec}^\top \textbf{X}_\text{rec},\\
        \mathbb{V}\text{ar}[\pi(x)] &:= \int | x -  \mathbb{E}[\pi(x)] |^2 \text{d} \pi(x) \approx \textbf{w}_\text{rec}^\top \text{diag} \Big[ (\textbf{X}_\text{rec} - \mathbb{E}[\pi(x)])^\top (\textbf{X}_\text{rec} - \mathbb{E}[\pi(x)]) \Big], \\
        |x^*_\text{true} - \mathbb{E}[\pi(x)]| &\approx \sqrt{\sum_{k=1}^d (x^*_{\text{true}, k} - (\textbf{w}_\text{rec}^\top \textbf{X}_\text{rec})_k)^2}, \\
        \text{BR} &:= | y^*_\text{true} - \textbf{w}_\text{batch}^\top f_\text{true}(\textbf{X}_\text{batch}) |.
    \end{align}
\end{subequations}
\gls{mv}, and \gls{md} corresponds to the $\pi$ shrinkage, of which smaller value indicates shrinking. \gls{md} represents the Euclidean distance between the mean of $\pi$ and the true global maximum $x^*_\text{true}$.
We compared these two metrics against \gls{br}. \gls{br} is the batch estimation regret (referred as \gls{br} for Theorem 2 in \cite{kandasamy2018parallelised}).
Experiments were done using \gls{sober}-\gls{lfi} on the Ackley function (see Table \ref{tab:exp}) over 10 iterations with 16 repeats 
(160 data points). Figure \ref{fig:corr} shows the linear correlation matrix of these 3 metrics. Both \gls{md} and \gls{mv} are highly correlated with \gls{br}, clearly explaining the $\pi$ shrinkage as the dual objective in Eq. \eqref{eq:dual} is the good measure of \gls{br}. 
In other words, $\pi$ (\gls{mc} estimate of $x^*$) shrinks toward true global maximum $x^*_\text{true}$ with being smaller variance (more confident), and both linearly correlated to minimising the Bayesian regret, \gls{br}.

\subsubsection{Hyperparameter sensitivity}
\begin{figure*}
  \centering
  \includegraphics[width=1.00\textwidth]{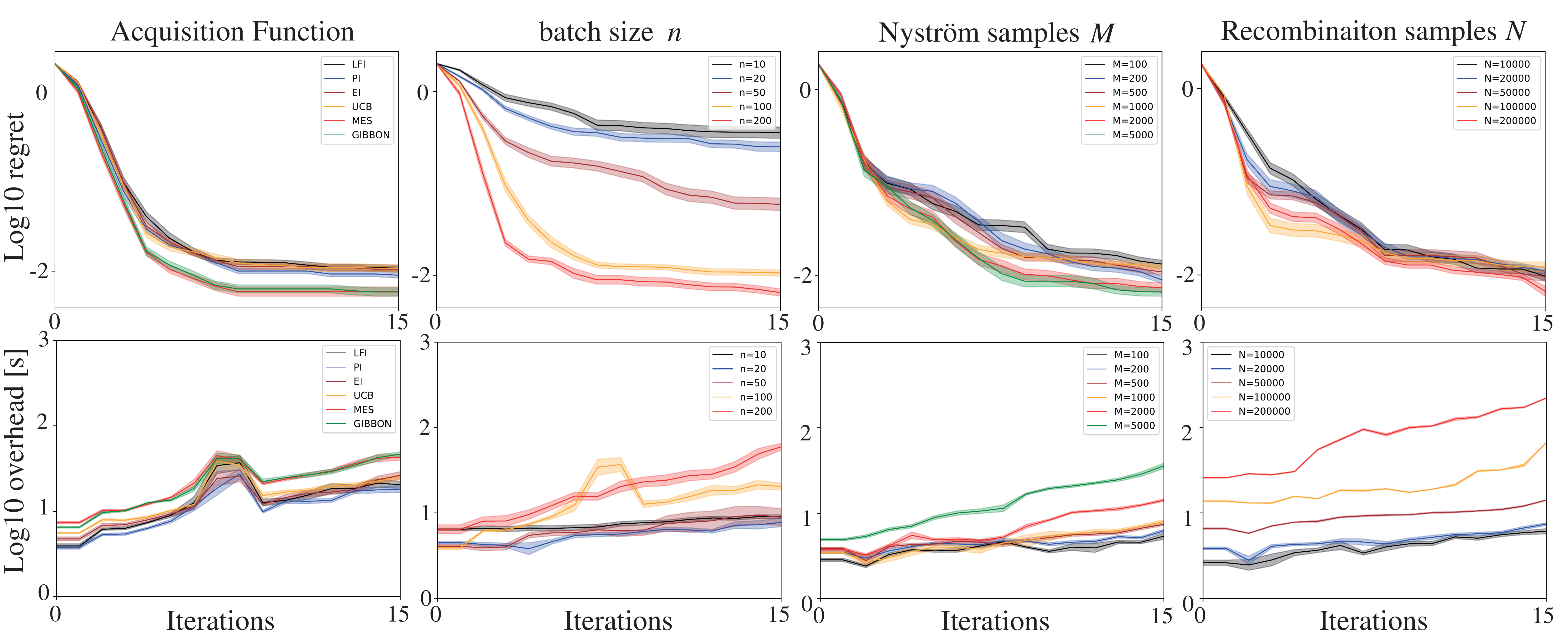}
  \caption{Hyperparameter sensitivity analysis using the Ackley function. Lines and shade area denote mean $\pm$ 1 standard error.}
  \label{fig:figs2}
\end{figure*}

We tested the hyperparameter sensitivity of \gls{sober}-\gls{lfi} using the Ackley function. We examined the effect of \glspl{af} $\alpha$, batch size $n$, the number of Nystr{\"o}m samples $M$, and the number of recombination samples $N$. We averaged the results from 16 experiments with varied random seeds, and terminated at the 15th batch acquisition. The baseline conditions are $n = 100$, $\alpha = \text{LFI}$, $M = 500$, and $N = 20,000$. For \gls{af}, the information-theoretic \glspl{af} can boost the convergence rate, whereas the others do not change significantly. For the batch size $n$, the convergence rate can be improved in accordance with the batch size. For quadrature hyperparameters $M$ and $N$, the larger the number of samples becomes, the faster the convergence does. However, increasing the number of samples leads to additional overhead increase. Our default conditions are competitive throughout the experiments.

\subsubsection{Ablation Study}
\begin{figure*}
  \centering
  \includegraphics[width=0.55\textwidth]{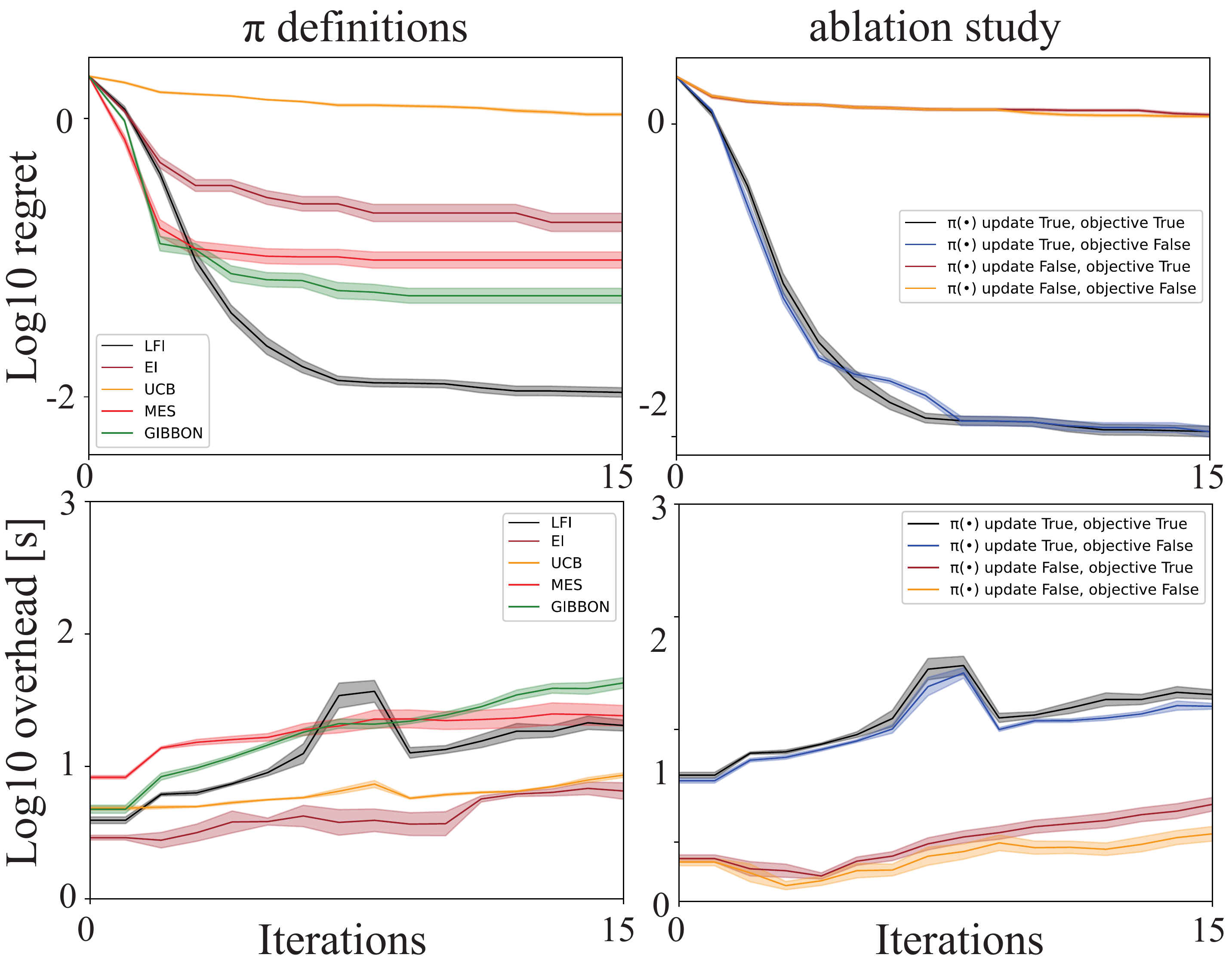}
  \caption{Ablation study using the Ackley function. Lines and shade area denote mean $\pm$ 1 standard error.}
  \label{fig:figs3}
\end{figure*}

We performed the ablation study to analyse each algorithm's effects on convergence rate. Firstly, we compared the various $\pi$ definitions defined by the \glspl{af}. As shown in Figure \ref{fig:figs3}, LFI and PI definitions are the clear performants. This is because the other \glspl{af} are designed to guide the sequential sampling, of which global maxima sensitively changed over iterations. LFI and PI show the possibility of global maxima, which gradually squeezes the region toward the true global maxima. Thus, in \gls{sober}-\gls{lfi} formulation, LFI \gls{af} is well-suited as the definition of $\pi$. As another ablation study, we compared whether or not updating $\pi$ and using \gls{af} in the objective \gls{rchq}. As a result, unsurprisingly, updating $\pi$ is the most influential on the convergence rate. The objective \gls{rchq} does not significantly influence the convergence when we select LFI as the objective. However, information-theoretic \glspl{af} can boost the convergence, as shown in Figure \ref{fig:figs2}.

\subsubsection{Fully Bayesian Gaussian Process}
\begin{figure*}
  \centering
  \includegraphics[width=0.7\textwidth]{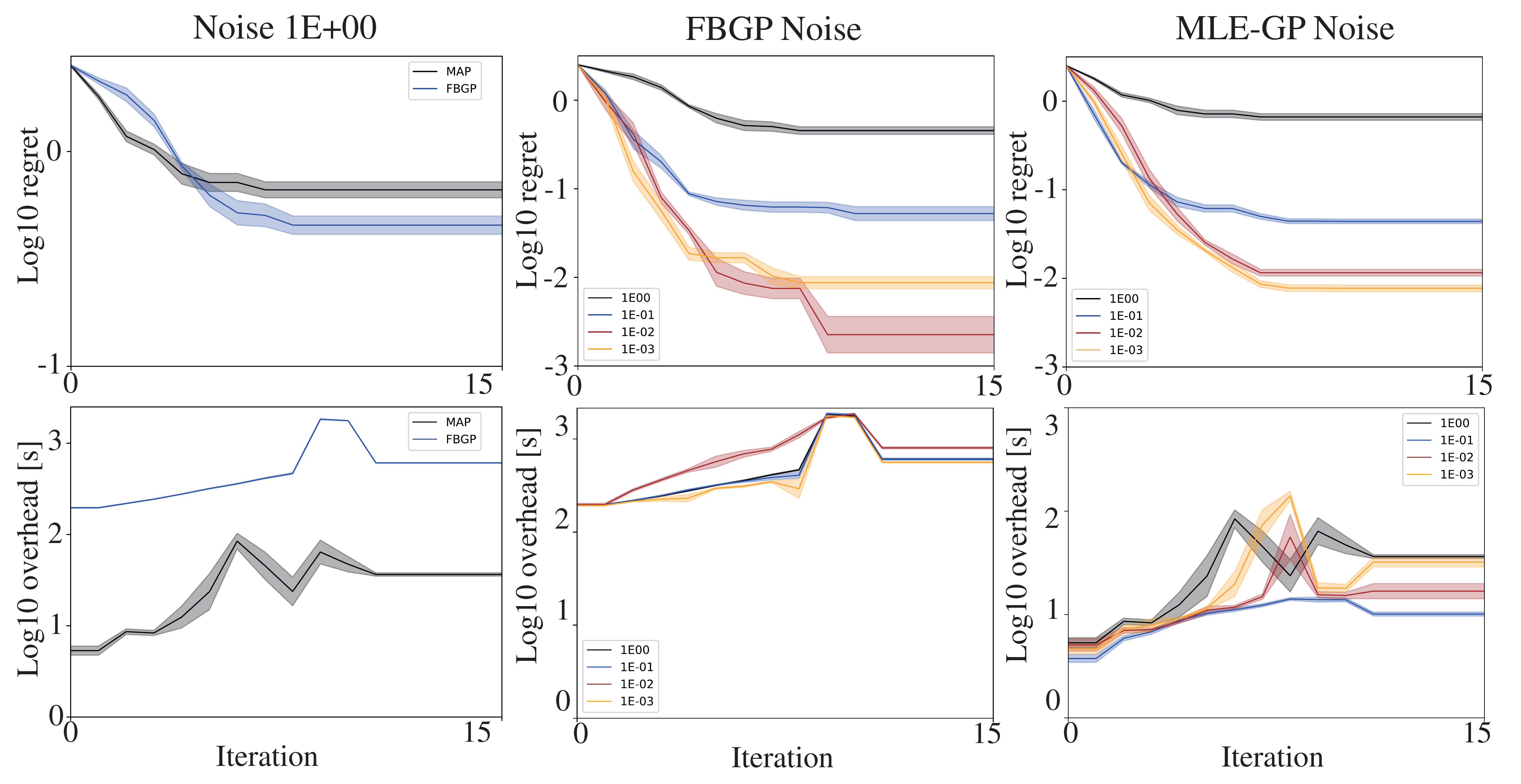}
  \caption{Efficacy of Fully Bayesian Gaussian process modelling using the noisy Ackley function. Lines and shade area denote mean $\pm$ 1 standard error.}
  \label{fig:figs4}
\end{figure*}

We further tested the effect of \gls{fbgp} modelling on the convergence rate. To examine the efficacy, we adopted the noisy Ackley function. We added the Gaussian noise to the queried values from the Ackley function. The amplitude of the noise is varied from $10^{-3}$ to $1$ in a logarithmic order. The baseline conditions are $n = 100$, $\alpha = \text{LFI}$, $M = 500$, $N = 20,000$, and $H = 50$. Figure \ref{fig:figs4} illustrates that \gls{fbgp} modelling with quadrature distillation can boost the convergence rate while maintaining the overhead feasibly small (the overhead of \gls{fbgp} is smaller than DPP-TS with type-II MLE kernel.)

\begin{figure*}
  \centering
  \includegraphics[width=0.55\textwidth]{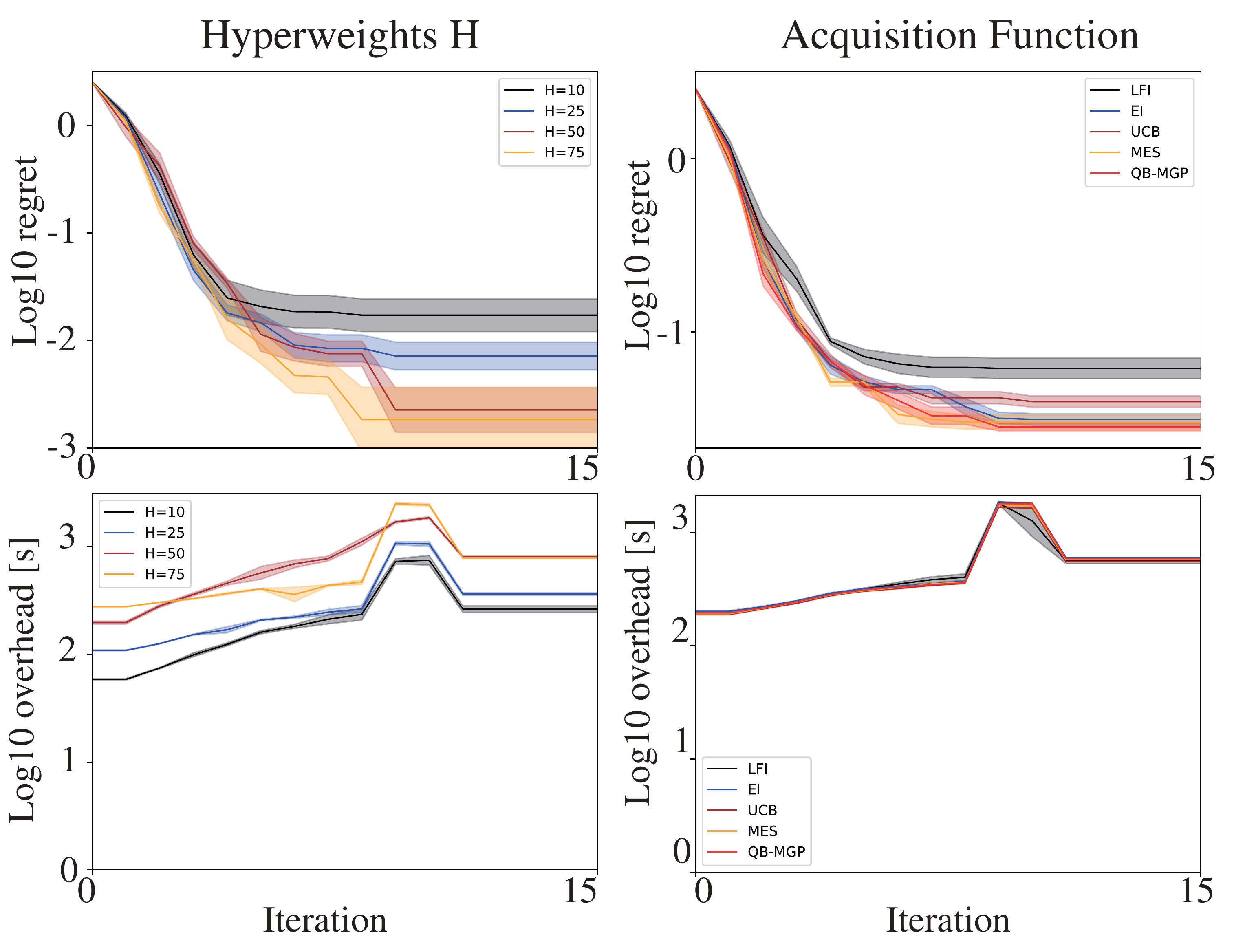}
  \caption{\gls{sober}-\gls{lfi} consistently outperforms with small overhead. Lines and shade area denote mean $\pm$ 1 standard error.}
  \label{fig:figs5}
\end{figure*}

Furthermore, we examined the effects of the number of hyperweights $H$ and the \glspl{af}. While $H$ are influential on both convergence rate and overhead, the default value $H=50$ are reasonably competitive. With regard to the effect of \glspl{af}, QB-MGP \gls{af} was the performant.

\end{document}